\documentclass{article}

    \usepackage[preprint]{main}

\usepackage[utf8]{inputenc} % allow utf-8 input
\usepackage[T1]{fontenc}    % use 8-bit T1 fonts
\usepackage{url}            % simple URL typesetting
\usepackage{booktabs}       % professional-quality tables
\usepackage{amsfonts}       % blackboard math symbols
\usepackage{nicefrac}       % compact symbols for 1/2, etc.
\usepackage{microtype}      % microtypography
\usepackage{xcolor}         % colors

\usepackage{graphicx}

\usepackage{graphicx}
\usepackage{wrapfig}
\usepackage{float}
\usepackage{xspace}
\usepackage{multirow}
\usepackage{colortbl}

\usepackage{caption}
\usepackage{subcaption}

\usepackage[utf8]{inputenc} % allow utf-8 input
\usepackage[T1]{fontenc}    % use 8-bit T1 fonts
\usepackage{url}            % simple URL typesetting
\usepackage{booktabs}       % professional-quality tables
\usepackage{amsfonts}       % blackboard math symbols
\usepackage{nicefrac}       % compact symbols for 1/2, etc.
\usepackage{microtype}

\usepackage{graphicx}

\usepackage{bbding}
\usepackage{amssymb}
\usepackage{graphicx}
\usepackage{wrapfig}
\usepackage{float}
\usepackage{xspace}
\usepackage{multirow}
\usepackage{colortbl}
\usepackage{makecell}
\usepackage[colorlinks=true       
            ]{hyperref}
\usepackage{pifont}
\usepackage{tcolorbox}
\usepackage[utf8]{inputenc} % allow utf-8 input
\usepackage[T1]{fontenc}    % use 8-bit T1 fonts
\usepackage{url}            % simple URL typesetting
\usepackage{booktabs}       % professional-quality tables
\usepackage{amsfonts}       % blackboard math symbols
\usepackage{nicefrac}       % compact symbols for 1/2, etc.
\usepackage{microtype}      
\usepackage{xcolor}         % colors

\usepackage{graphicx}

\usepackage{bbding}
\usepackage{amssymb}
\usepackage{graphicx}
\usepackage{wrapfig}
\usepackage{float}
\usepackage{xspace}
\usepackage{multirow}
\usepackage{colortbl}
\usepackage{makecell}
\usepackage{caption}
\usepackage{subcaption}
\usepackage{pifont}
\usepackage{tcolorbox}
\makeatletter
\DeclareRobustCommand\onedot{\futurelet\@let@token\@onedot}
\def\@onedot{\ifx\@let@token.\else.\null\fi\xspace}

\def\eg{\emph{e.g}\onedot} 
\def\ie{\emph{i.e}\onedot} 
 
\def\etc{\emph{etc}\onedot}

\makeatother
\definecolor{green1}{rgb}{0.13, 0.55, 0.13}
\makeatletter
\DeclareRobustCommand\onedot{\futurelet\@let@token\@onedot}
\def\@onedot{\ifx\@let@token.\else.\null\fi\xspace}

\def\eg{\emph{e.g}\onedot} 
\def\ie{\emph{i.e}\onedot} 
 
\def\etc{\emph{etc}\onedot}

\makeatother

\title{ChatBridge: Bridging Modalities with \\
Large Language Model as a Language Catalyst}

\author{%
  Zijia Zhao$^{1,3}$ , Longteng Guo$^2$ , Tongtian Yue$^{1,3}$, \\\textbf{Sihan Chen}$^{1,3}$, \textbf{Shuai Shao}$^2$, \textbf{Xinxin Zhu}$^{1,3}$, \textbf{Zehuan Yuan}$^2$ , \textbf{Jing Liu}$^{1,3}$ \\
  $^{1}$Institute of Automation, Chinese Academy of Sciences  \qquad $^{2}$Bytedance Inc. \\
  $^{3}$School of Artificial Intelligence, University of Chinese Academy of Sciences \\
  \url{https://iva-chatbridge.github.io}
}

\begin{document}

\maketitle

\begin{abstract}
Building general-purpose models that can perceive diverse real-world modalities and solve various tasks is an appealing target in artificial intelligence. 
In this paper, we present ChatBridge, a novel multimodal language model that leverages the expressive capabilities of language as the catalyst to bridge the gap between various modalities. We show that only language-paired two-modality data is sufficient to connect all modalities. 
ChatBridge leverages recent large language models (LLM) and extends their zero-shot capabilities to incorporate diverse multimodal inputs. 
ChatBridge undergoes a two-stage training. The first stage aligns each modality with language, which brings emergent multimodal correlation and collaboration abilities. The second stage instruction-finetunes ChatBridge to align it with user intent with our newly proposed multimodal instruction tuning dataset, named MULTIS, which covers a wide range of 16 multimodal tasks of text, image, video, and audio modalities. 
We show strong quantitative and qualitative results on zero-shot multimodal tasks covering text, image, video, and audio modalities. 
All codes, data, and models of ChatBridge will be open-sourced.
\end{abstract}

\section{Introduction}
Humans interact with the world through multiple modalities — we see objects, hear sounds, feel textures, smell odors, speak words, and so on. By leveraging complementary information from each modality, we obtain a comprehensive understanding of our surroundings. 

In order for Artificial Intelligence to complete various real-world tasks in the wild, it needs to be able to interpret, relate, and reason about information from multiple modalities. Significant processes have been made in multimodal learning applications, including vision and language learning~\cite{uniter,visualbert,anderson2018bottom,albef}, video understanding~\cite{valor,xclip}, audio-visual speech recognition~\cite{ma2021avsroc}, autonomous driving \etc. However, current paradigms in multimodal learning often still require acquiring all types and combinations of paired data, and their capabilities are often limited to solving specific tasks with model tuning, \eg, visual question answering, sentiment analysis \etc. 

In this paper, we present \textit{ChatBridge}, a unified multimodal model that harnesses the power of advanced large language model (LLM) as a language catalyst to interpret, correlate, and reason about various modalities, and can perform zero-shot tasks of human instructions through multi-round dialogues. 
Large language models, such as ChatGPT~\cite{chatgpt}, GPT-4~\cite{gpt4}, and LLAMA~\cite{touvron2023llama}, have demonstrated exceptional proficiency in understanding and generating human-like text.
They show that language can act as a universal interface for a general-purpose assistant, where various tasks can be explicitly represented and responded to in language. 
By extending LLMs' capabilities to incorporate diverse multimodal inputs, we devise a multimodal language model that can perceive real-world modalities, as well as follow instructions, think, and interact with humans in natural language. 

Our method doesn't require datasets where all modalities co-occur with each other. Instead, we leverage language as the catalyst to bridge modalities -- we only require easy-acquired, language-paired two-modality data (\eg, image-text pairs, video-text pairs, audio-text pairs, \etc). Such strategy leads to an emergent multimodal correlation and collaboration across all of the modalities, enabling zero-shot perception capabilities on multimodal inputs without explicitly paired training data (\eg the rare video-audio-text triples data where the text description describes both the video and audio contents).

Specifically, ChatBridge integrates multiple modality-specific encoders and an LLM, Vicuna~\cite{vicuna2023}, which is built upon LLaMA~\cite{touvron2023llama}, with learnable perceiver modules in between to project embeddings from different modalities into the semantic space of LLM. 
ChatBridge undergoes a two-stage training on large-scale language-paired two-modality data and self-built multimodal instruction-following data.
In the first stage, we pretrain ChatBridge to align each modality with language, which brings emergent multimodal correlation and collaboration abilities with LLM as a language catalyst. In the second stage, we instruction-finetune ChatBridge to align the model with user intent on our newly collected MULTimodal InStruction tuning dataset (MULTIS), enabling more effective zero-shot generalization on multimodal tasks.
MULTIS covers a wide range of 16 multimodal task categories and 15 source datasets involving image, video, and audio content. It consists of both standardized task-specific data and open-ended multimodal chat data.

\begin{wrapfigure}{rt!}{0.4\textwidth}
    \centering
    \vspace{-\baselineskip}\setlength\intextsep{0pt}
    \includegraphics[width=0.3\textwidth]{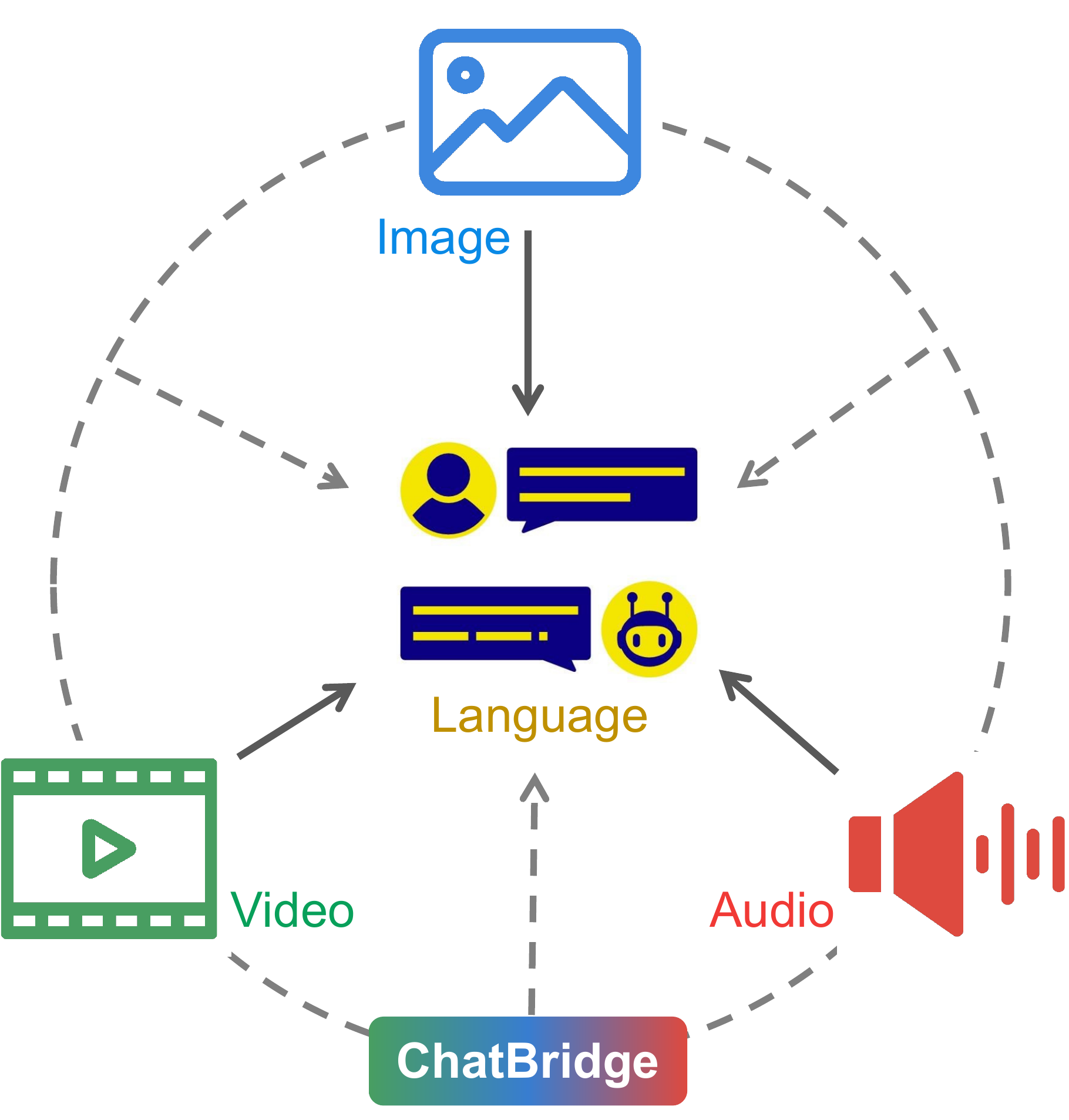}
  \caption{ChatBridge overview. We harness the power of advanced LLM as the catalyst to bridge modalities with easy-acquired, language-paired two-modality data (\eg, image-text, video-text, and audio-text), resulting in a multimodal LLM with emergent multimodal correlation and collaboration abilities across all of the modalities. }
  \vspace{-1cm}
\end{wrapfigure}

Our paper makes the following contributions:
\begin{itemize}
    \item We propose ChatBridge, an approach to learning a unified multimodal model to interpret, correlate, and reason about various modalities without relying on all combinations of paired data.
    \item We collect MULTIS, a multimodal instruction tuning dataset that consists of 16 diverse multimodal tasks covering text, image, video, and audio modalities.
    \item We quantitatively and qualitatively evaluate ChatBridge on a wide range of zero-shot multimodal tasks covering text, image, video, and audio modalities, and demonstrate that ChatBridge achieves strong zero-shot performance on these tasks.
    \item We will open-source the codebase, the MULTIS data, the model checkpoint, and a multimodal chat demo to facilitate future research toward building a general-purpose multimodal assistant.
\end{itemize}

\section{Related works}
\subsection{Multimodal Learning}
Multimodal learning aims to acquire knowledge from diverse forms of data and apply it to various tasks involving multiple modalities. Some approaches typically employ joint learning strategies and generate a merged embedding as the complementary multimodal representation across two modalities, including vision-language~\cite{visualbert, uniter, albef, anderson2018bottom}, audio-visual~\cite{yang2022icode,avsdsota,musicqa}, RGB-D~\cite{wu2022trans-d, rgb-d, chen2020mmfnet}, and speech-text~\cite{zheng2021fused}, among others. 
In contrast, some other methods focus on learning separate representations for each modality and then aligning them in a shared latent space with metric learning. The seminal work CLIP~\cite{clip} initially maps visual and textual embeddings into the same space, achieving remarkable performance in retrieval and classification tasks. Subsequent methods~\cite{guzhov2022audioclip, xclip} have extended this paradigm to incorporate additional modalities. More recently, ImageBind~\cite{girdhar2023imagebind} has proposed a multimodal method for aligning different modalities with images, following a similar paradigm. 

ChatBridge projects embeddings from different modalities into the semantic space of LLM and facilitates modality collaboration within LLM. This integration harnesses the strengths of both multimodal learning methods, resulting in a more comprehensive and effective approach.
\subsection{LLM and Multimodal LLM}
In recent years, there has been remarkable progress in the development of LLMs, particularly in the context of GPT-3\cite{gpt3}. As a result, numerous LLMs have been introduced, such as OPT\cite{zhang2022opt}, BLOOM\cite{scao2022bloom}, PaLM\cite{chowdhery2022palm}, GLM\cite{zeng2022glm}, and LLaMA\cite{touvron2023llama}. 
The success of the general purpose assistant ChatGPT\cite{chatgpt} has inspired researchers to explore methods for emulating its proficiency by employing instruction tuning techniques on language models~\cite{instructgpt, selfinstruct, alpaca, vicuna2023,flant5} and vision-language models~\cite{llava, xu2022multiinstruct, minigpt4}.

Despite the successful applications of LLMs in processing language, their ability to comprehend modalities beyond text, such as vision and audio, remains a challenge. Recently, researchers have made efforts to extend language models to incorporate visual inputs, employing two distinct paradigms: systematic collaboration and end-to-end trained approaches.
Systematic collaboration approaches, exemplified by Visual ChatGPT~\cite{wu2023visualchatgpt}, MM-REACT~\cite{yang2023mmreact}, HuggingGPT~\cite{shen2023hugginggpt} and ChatVideo~\cite{wang2023chatvideo}, leverage various vision experts or tools to express visual information through textual descriptions. In these methods, LLMs, such as ChatGPT~\cite{chatgpt}, act as agents and are prompted to select the appropriate experts and tools for visual comprehension.
On the other hand, end-to-end trained approaches utilize LLMs to construct unified image-based multimodal models. Flamingo~\cite{flamingo} freezes the pretrained vision encoder and LLM, integrating visual and language modalities using gated cross-attention, which exhibits impressive few-shot capabilities. BLIP-2~\cite{blip2} employs Q-Former to align visual features from the frozen visual encoder and LLMs. Additionally, PaLM-E~\cite{driess2023palme} directly incorporates features from sensor modalities into PaLM~\cite{chowdhery2022palm}. There are also several methods built on open-sources LLM  LLaMA~\cite{touvron2023llama} and its variations Alpaca~\cite{alpaca} and Vicuna~\cite{vicuna2023}, including LLaVA~\cite{llava}, MiniGPT4~\cite{minigpt4} and mPLUG-Owl~\cite{ye2023mplugowl}.

\section{Methods}
ChatBridge is a multimodal language model capable of perceiving real-world multimodal information, as well as following instructions, thinking, and interacting with humans in natural language. In this paper, we consider the modalities of image, video, and audio, while deferring the integration of additional modalities such as sketch and point cloud to future works.

\subsection{Architecture Overview}
\begin{figure}
  \centering
     \includegraphics[width=0.9\linewidth]{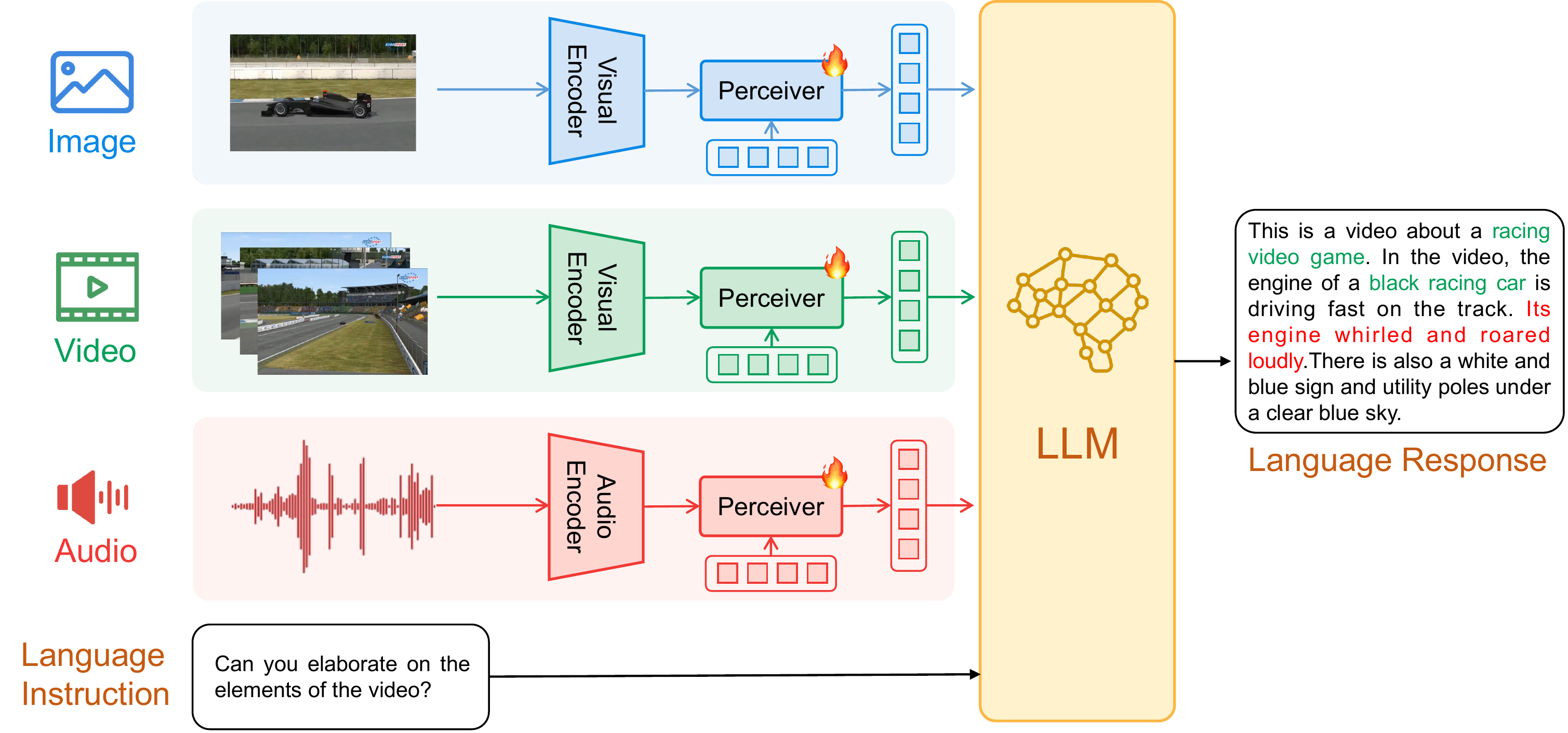}
  \caption{Model architecture of ChatBridge. It consists of multiple modal-specific encoders and perceiver modules and an LLM. }
  \vspace{-0.5cm}
\label{arch}
  
\end{figure}

As illustrated in Figure~\ref{arch}, ChatBridge consists of multiple modal-specific encoders and perceiver modules, and a transformer-decoder-based LLM. 
Inspired by Flamingo~\cite{flamingo} and BLIP-2~\cite{blip2}, we introduce perceiver modules to bridge the encoders and the LLM. 
The perceiver summarizes the variable-length embeddings from each encoder's outputs within a given number of learnable query tokens. It thereby produces outputs of the same shape for all modalities. Also, as the number of query tokens is much smaller than the size of encoder features, it significantly reduces the computation cost in LLM. We instantiate the perceiver as a transformer decoder with learnable query tokens and the encoder embeddings as the input. 

The information $\mathbf{X}_i$ from the $i$-th modality is first fed into the encoder $h_{i}$ to extract its features. And then each perceiver $\rho_i$ with learnable queries $\mathbf{H}_{i}$ transforms them into a shared latent space:
\begin{equation}
    \mathbf{Z}_{i} = \rho_i\left( Q=\mathbf{H}_{i}, K, V=h_i(\mathbf{X}_i) \right)
\end{equation}
Given the all multimodal inputs and human instruction $\mathbf{X}_{instruction}$ as inputs, the LLM $f$ generates the final response text sequence $\mathbf{Y}$ by:
\begin{equation}
    \mathbf{Y} = f \left( \mathbf{Z}_{1}, \mathbf{Z}_{2},...,\mathbf{Z}_{n}, \mathbf{X}_{instruction} \right)
\end{equation}

Specifically, we choose open-sourced Vicuna-13B~\cite{vicuna2023} as the LLM, which is built upon LLaMA, and reported to achieve $90\%$ of ChatGPT’s quality as per GPT-4’s evaluation. As for the modal-specific encoders, we choose ViT-G~\cite{evaclip} as the vision encoder to encode images and videos, and BEAT~\cite{beats} as the audio encoder to encoder audios. We sample 4 frames from each video and concatenate their respective frame features to form the video features, which are inputs of the video perceiver. Similarly, for each audio, we divide it into clips of 10-second intervals and concatenate the clip features to create the audio features. 
We use a shared perceiver for all modalities while each modality has its independent learnable query tokens. Due to limited computation resources, we only train the perceivers and their learnable query tokens while keeping the encoders and LLM frozen during the whole training process. 

\subsection{Two-stage Training}
Motivated by ChatGPT~\cite{chatgpt} that is built upon the pretrained GPT-3.5, ChatBridge also undergoes a two-stage training on large-scale language-paired two-modality data, and self-built multimodal instruction-following data. 

\paragraph{Stage1: Multimodal Alignment}
In the first stage, we pretrain ChatBridge to align each modality with language, which brings emergent multimodal correlation and collaboration abilities with LLM as a language catalyst. We leverage large-scale language-paired two-modality data for multimodal alignment training, including image-text, video-text, and audio-text pairs.   
Specifically, the training data consists of publicly available datasets of image-text pairs (including MS-COCO~\cite{coco}, SBU Captions~\cite{sbu}, Conceptual Captions~\cite{cc3m, cc12m}, LAION-115M~\cite{laion400m}), video-text pairs of Webvid10M~\cite{webvid10m}, and audio-text pairs of WavCaps~\cite{mei2023wavcaps}. The raw unimodal data (\ie images, videos, audios) is sequentially fed into the modality-specific encoder and perceiver to get unimodal embeddings. 
The input format of LLM in this training stage is:
\begin{center}
``\emph{<unimodal input><text>}"
\end{center}
where \emph{<unimodal input>} is the sequence of unimodal embeddings from the perceiver, which can be regarded as soft prompts. The LLM directly tasks \emph{<unimodal input>} as input and is trained to output the corresponding text \emph{<text>} in the training samples.
We train for 150k steps in this training stage with a batch size of 256 on 8 A100 GPUs.

\paragraph{Stage2: Multimodal Instruction Tuning}
After aligning unimodal data with LLM, our model already has the basic ability to understand information from various modalities. However, the model still needs to improve its ability for processing different modalities and following human instructions. 
Some previous methods~\cite{instructgpt, selfinstruct, alpaca, vicuna2023} have proved that tuning the large model with instructions can help it to understand the intent of human beings. Inspired by these methods, in the second training stage, we further instruction-finetune ChatBridge to align the model with user intent on a wide range of multimodal tasks, enabling more effective zero-shot generalization on multimodal tasks. 
To this end, we carefully collect a multimodal instruction tuning dataset to funetune our model, where the instructions are multimodal containing text, image, video, and audio, while the responses are text only. A specific introduction to the data collection process will be provided in Section~\ref{datagen}. We organized all the samples in a standardized format as below:

\begin{center}
``\emph{\#\#\#Human: <multimodal input prompt>, <instruction>\#\#\#Assistant: <response>}"
\end{center}
where \emph{<multimodal input prompt>} is some human-craft template prompt that combines multiple sequences of unimodal embeddings from the perceivers. The LLM ingests the whole sequence and is trained to output the correct response \emph{<response>}. This training stage costs 10k steps with a batch size of 4k tokens on 8 A100 GPUs.

\subsection{Multimodal Instruction Tuning Dataset --- MULTIS}
\label{datagen}
\begin{figure}[b]
  \centering
  \vspace{-0.5cm}
     \includegraphics[width=\linewidth]{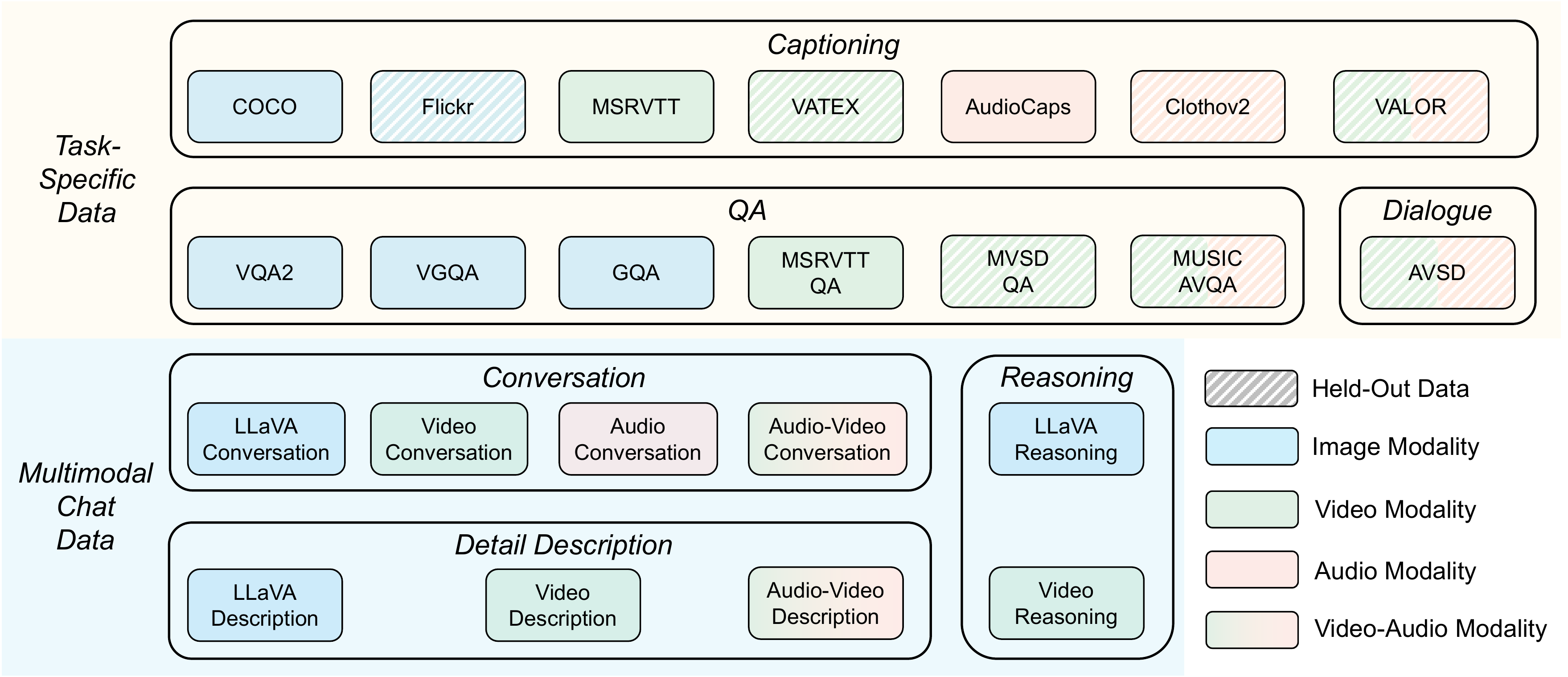}
    \vspace{-0.5cm}
  \caption{Tasks groups included in MULTIS multimodal instruction tuning dataset.}
  \label{fig:multis}
    % \vspace{-0.5cm}
\end{figure}

We have developed a diverse dataset for multimodal instruction-tuning, named MULTIS, to instruction-finetune ChatBridge model. MULTIS consists of two distinct parts: task-specific data and multimodal chat data. The former presents standardized tasks that require concise responses, while the latter simulates real-world problem-solving scenarios by featuring open-ended dialogue between a human and a multimodal assistant. As shown in Figure~\ref{fig:multis}, the whole collection of MULTIS covers 16 multimodal task categories and 15 source datasets. We hold out 6 datasets for model evaluation purposes. 

\subsubsection{Task-Specific Data}
We collect a vast array of publicly available human-annotated multimodal datasets and transform them into a unified instruction tuning format. Specifically, a plethora of common Question-Answering (QA) and captioning datasets that contain image-text, video-text, and audio-text pairs are assembled, encompassing VQAv2\cite{vqa}, VG-QA\cite{vg}, COCO Caption\cite{coco}, MSRVTTQA\cite{msrvtt}, MSRVTT Caption\cite{msrvtt}, AudioCaps\cite{audiocaps}. For each task, we employ ChatGPT~\cite{chatgpt} to derive 10\textasciitilde15 unique instruction templates, which are then manually filtered and refined to ensure rationales and diversity are optimal. As the public datasets inherently favor shorter responses, we craft instruction template modifiers to specify the desired response style, such as \textit{short} and \textit{brief} for short-answer data, and \textit{a sentence} and \textit{single sentence} for caption data. 

\subsubsection{Multimodal Chat Data}
While task-specific data empowers the model towards completing standardized tasks, multimodal chat data offers real-world, open-ended dialogues demanding more sophisticated intent comprehension and contextual reasoning abilities, as well as providing more diverse, helpful, human-like responses. Despite the image-to-text chat dataset generated by LLaVA-Instruct-150K~\cite{llava}, chat data across other modalities remains limited. To this end, we have constructed a multimodal chat dataset that comprises both unimodal and multimodal inputs of image, video, and audio modalities.

\begin{figure}
  \centering
     \includegraphics[width=\linewidth]{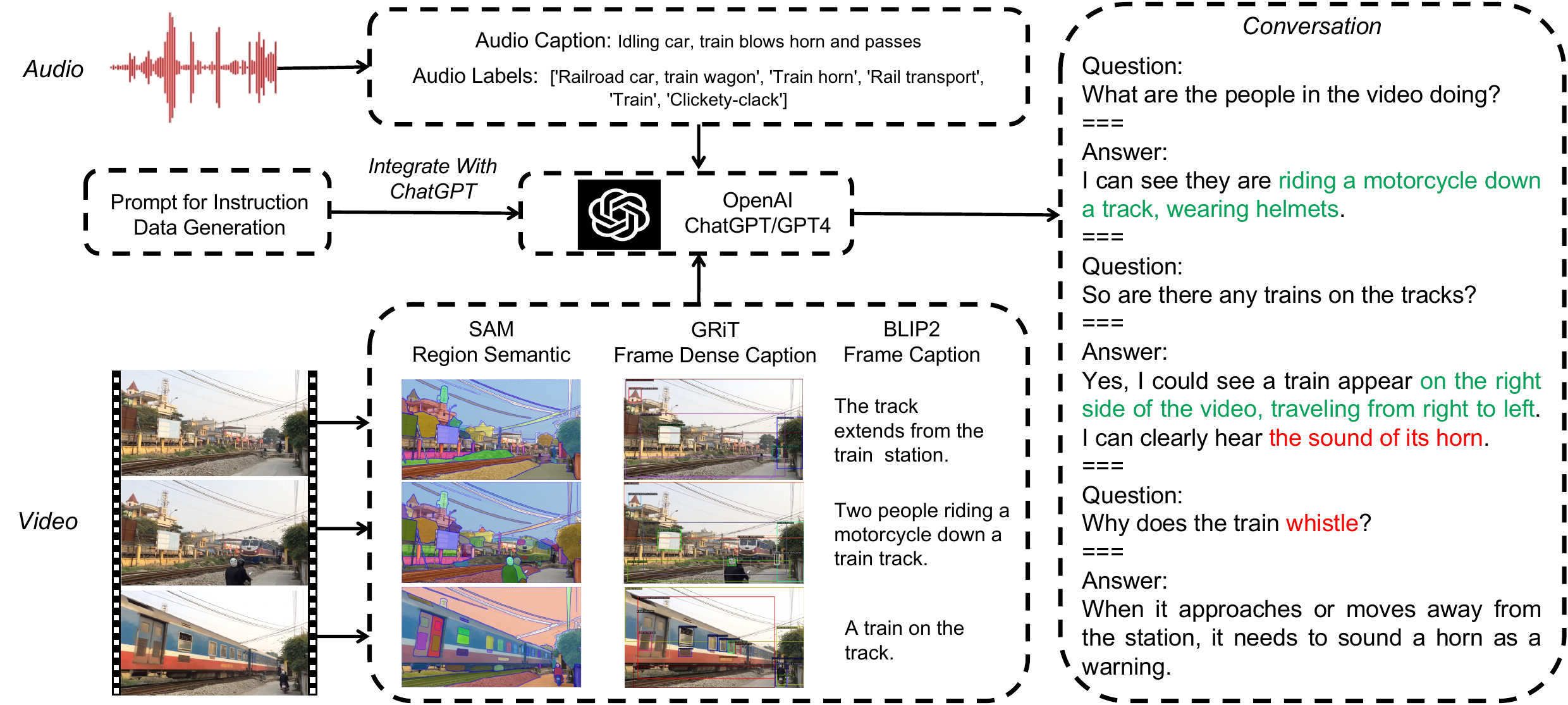}
     \vspace{-0.3cm}
  \caption{Illustration of the pipeline to collect the multimodal chat data for MULTIS.}
  \vspace{-0.5cm}
\label{fig:datagen}
\end{figure}

We adopt LLaVA-Instruct-150K~\cite{llava} as the image chat data. To incorporate additional modalities, namely video, audio, and video-audio content, we design a pipeline with the assistance of offline open-source models\cite{sam, semanticsam, grit, blip2, Image2Paragraph} and ChatGPT\cite{chatgpt}, as is shown in Figure~\ref{fig:datagen}. Following~\cite{llava}, we generate three types of instruction-following data including conversation, detailed description, and complex reasoning. We obtain our video from the MSRVTT~\cite{msrvtt} dataset, audio from AudioCaps~\cite{audiocaps}, and video-audio from VALOR~\cite{valor}. 
In order to prompt a text-only ChatGPT or GPT-4, we convert those non-textual modalities into textual descriptions.
Specifically, for each video, we extract three frames and employ Semantic-SAM\cite{semanticsam}, GRIT\cite{grit}, and BLIP-2~\cite{blip2} to develop annotations of the frames at the region semantic, region caption, and frame caption levels. We then concatenate these frame annotations in temporal order with the human-annotated video captions utilizing hand-crafted prompts. For each audio, we use its audio captions and labels from the original dataset. We combine those collections of fine-grained and global descriptions with manually designed seed examples to query ChatGPT or GPT-4 in an in-context-learning manner. With the above pipeline, we collect video, audio, and video-audio multimodal chat samples comprising 24k in conversations, 18K in detailed descriptions, and 9k in complex reasoning.

Overall, MULTIS contains 4.4M task-specific samples and 209k multimodal chat samples.

\section{Experiments}
\subsection{Zero-shot Task Evaluation}

We evaluate ChatBridge's zero-shot ability on the held-out datasets of MULTIS's task-specific data. The model is instructed with \emph{unimodal} and \emph{multimodal} inputs along with text instructions to generate the corresponding answers. \footnote{Please note that here "unimodal input" and "multimodal input" refer to the use of text and either a single or multiple modalities in image, video, and audio. }

\paragraph{Unimodal Input Tasks} 
Unimodal input tasks comprise of question answering (QA) and captioning tasks on image-text (OKVQA~\cite{okvqa}, GQA~\cite{gqa}, Flickr Captioning~\cite{flickr},  nocaps~\cite{agrawal2019nocaps}), video-text (MSVDQA~\cite{msrvtt}, VATEX~\cite{vatex}), and audio-text (clothoV2~\cite{drossos2020clotho}) datasets. 
QA tasks require the model to predict a short answer about the unimodal input, while captioning tasks require outputting a sentence description. 
As shown in Table~\ref{table-unimodal}, ChatBridge exhibits remarkable performance on unimodal input tasks, indicating successful alignment of unimodal input and language. 
On image-text datasets, our method achieves comparable performance as advanced image-based methods, Flamingo~\cite{flamingo} and BLIP-2~\cite{blip2}, and  
achieves new zero-shot state-of-the-art (SoTA) on Flickr30k and VATEX captioning tasks.

\begin{table}[t]
\caption{Zero-shot evaluation of SoTA methods on unimodal input tasks. We report the accuracy for QA tasks and the CIDEr~\cite{cider} score for captioning tasks. }
        \label{table-unimodal}
    \centering

 \resizebox{\linewidth}{!}{

\begin{tabular}{l|cccc|cc|c}
\toprule
\multirow{3}{*}{Methods}  & \multicolumn{4}{c|}{Image-Text Tasks}                    & \multicolumn{2}{c|}{Video-Text Tasks} & \multicolumn{1}{c}{Audio-Text Tasks} \\
     & OKVQA & GQA  & Flickr30k & NoCaps & MSVD & VATEX & Clothov2      \\
   & QA  &  QA  & Caption & Caption  & QA & Caption & Caption     \\   
 \midrule \rowcolor{gray!40} 
Finetuned SoTA &
 66.1~\cite{driess2023palme}
   & 65.1~\cite{zhao2021proto}
   & 67.4~\cite{unifiedVLP}
   & 121.6~\cite{blip2}
   
   & 60.0~\cite{valor}
   & 95.8~\cite{valor}
   & 48.8~\cite{mei2023wavcaps}

   \\ 
Flamingo-9B~\cite{flamingo}        & 44.7  &   -   & 61.5      &   -        & 30.2   &   39.5     & -                \\
Flamingo-80B~\cite{flamingo}            & \textbf{50.6}  &    -  & 67.2      &    -       & 35.6   &    46.7    & -                \\
BLIP-2 (FlanT5-XXL)~\cite{dai2023instructblip}     & -  & \textbf{42.4} & 73.7      & 98.4     & 34.4   & -   & -             \\
BLIP-2 (Vicuna-13B)~\cite{dai2023instructblip}      &  -     & 32.3 & 71.6      & 103.9    & 20.3   & -   & -                  \\  
ChatBridge w/o MULTIS             & 41.4  &  37.4    & 77.7      & 107.5     & 23.5   &    47.7    & 22.4               \\ 
ChatBridge            & 45.2  &  41.8    & \textbf{82.5}      & \textbf{115.7}   & \textbf{45.3}   &   \textbf{48.9}    & \textbf{26.2}           \\
 \bottomrule    
\end{tabular}

}
\vspace{-0.5cm}
\end{table}

\begin{table}[t]
\caption{Zero-shot evaluation of the effect of multimodal inputs on multimodal input tasks.}
        \label{table-multimodal}
    \centering

 \resizebox{0.7\linewidth}{!}{
\begin{tabular}{l|cc|cc|c}
\toprule
 \multirow{2}{*}{Input Modality}     & \multicolumn{2}{c|}{AVSD Dialogue}        & \multicolumn{2}{c|}{VALOR Captioning}   & MUSIC-AVQA \\
     & BLEU-4    & CIDEr   & BLEU-4   & CIDEr    & Acc.     \\ 
 \midrule \rowcolor{gray!40} 
Finetuned SoTA & 40.0~\cite{avsdsota}  & 108.5~\cite{avsdsota} & 9.6~\cite{valor} & 61.5~\cite{valor} &  78.9~\cite{valor}  \\ 
Video & 28.3 & 73.1 & 2.8 & 22.3 & 33.1   \\
Audio & 20.2 & 46.2 & 0.3 & 5.2  & 28.9   \\
Video+Audio & \textbf{29.8}                 & \textbf{75.4} & \textbf{4.2}                     & \textbf{24.7} & \textbf{43.0}                \\
\bottomrule
\end{tabular}}
\vspace{-0.5cm}
\end{table}

\begin{table}[t]
    \caption{Multimodal chat evaluation results. Response qualities of different methods are assessed by GPT-4 (text-only).} 
    \vspace{-0.1cm}
    \label{Tab}
    \begin{subtable}[t]{.47\linewidth}
    \caption{Multimodal chat evaluation on \textbf{image-text} chat data. }
    \vspace{-0.1cm}
    \label{table-unichat}
    \centering
     \resizebox{\linewidth}{!}{
\begin{tabular}{l|ccc|c}
\toprule
\multirow{2}{*}{Methods}  & \multirow{1}{*}{Reason-}  & Descrip-   & \multirow{1}{*}{Conver-}  & \multirow{2}{*}{Overall} \\
& -ing & -tion   &  -sation &  \\
\midrule
LLaVA~\cite{llava}    &    \textbf{8.87}       &         7.07           &     \textbf{7.63}         &  \textbf{7.86}       \\
BLIP-2~\cite{blip2}   &    5.80       &       6.00             &       7.03       &   6.28      \\
MiniGPT-4~\cite{minigpt4} &     7.53      &       \textbf{7.27}             &      5.63        & 6.14        \\
ChatBridge     &     7.17      &         6.23           &      7.23        & 6.88    
\\ \bottomrule
\end{tabular}}
    \end{subtable} \hspace{.05in}
    \begin{subtable}[t]{.47\linewidth}
      \caption{Multimodal chat evaluation on \textbf{video-audio-text} chat data. }
      \vspace{-0.1cm}
\label{table-multichat}
    \centering
     \resizebox{\linewidth}{!}{
\begin{tabular}{l|ccc|c}
\toprule
Input  & \multirow{1}{*}{Underst-}  & Reason-   & \multirow{1}{*}{Know-}  & \multirow{2}{*}{Overall} \\
Modality    & -anding &  -ing  & -ledge &  \\ \midrule
Video    &   5.86      &        5.27         &    7.70         &    6.15   \\
Audio   &   2.43     &       3.77         &    8.09         &   4.24   \\
Video+Audio &  \textbf{6.10}    &      \textbf{6.73}         &    \textbf{8.43}       &   \textbf{6.87}    \\
 \bottomrule
\end{tabular}}
    \end{subtable}
% \vspace{-0.4cm}
\end{table}

\paragraph{Multimodal Input Tasks} Multimodal input tasks necessitate the ability in interpreting, correlating, and reasoning about cross-modal information. 
We evaluate on multimodal input tasks encompassing 
audio-visual question answering (MUSIC-AVQA~\cite{musicqa}), audio-visual dialogue (AVSD~\cite{avsd}), and audio-visual captioning (VALOR~\cite{valor}) tasks. 
These tasks analyze videos containing both visual and auditory content.
Since our model represents a pioneering approach in handling multi-modal inputs, we perform ablation on the input modalities, as shown in Table~\ref{table-multimodal}.
Our model achieves better performance across all three tasks when incorporating both video and audio for solving these tasks, validating its capability to correlate and cooperate different modalities. 
We observe that video information has a greater influence on video-audio tasks, resulting in relatively higher performance for the video-only input modality as opposed to the audio-only counterpart. Nonetheless, amalgamating audio and video details enhances performance across all tasks to varying extents. Consequently, there is potential for further exploration in constructing a modality-balanced multimodal evaluation benchmark.

\paragraph{Effect of Instruction Tuning with MULTIS Data}
We also conduct a comparison of the performance in zero-shot task evaluation before and after applying our multimodal instruction tuning training stage.
As depicted in Table~\ref{table-unimodal}, after instruction tuning, the model demonstrates varying degrees of performance improvement across different downstream tasks. Specifically, we observe a 21.8\% boost in accuracy in MSVDQA, a 3.8\% improvement in OKVQA, and a 3.6\% improvement in GQA. 
Moreover, our approach also yields advancements in captioning tasks concerning Cider Score.
The experimental results demonstrate that the utilization of multimodal instruction tuning data within MULTIS can facilitate the model's abilities in integrating diverse modalities and generalizing effectively to unseen tasks.

\subsection{Multimodal Chat Evaluation}
\paragraph{Chat with Unimodal Input}
We conduct a comparative analysis of our model with three image-based LLMs: BLIP-2\cite{blip2}, LLaVA\cite{llava}, and MiniGPT-4\cite{minigpt4}. The evaluation is based on GPT-4 generated image-text chat data provided by LLaVA\cite{llava}, comprising 90 samples. We follow the evaluation protocol proposed by LLaVA\cite{llava}, where GPT-4 is used to evaluate the helpfulness, relevance, accuracy, and level of detail of each model’s responses. The responses are scored on a scale of 1 to 10, where a higher score indicates superior overall performance. 
As depicted in Table~\ref{table-unichat}, our method surpasses other approaches that also employ a frozen LLM, \ie BLIP-2 and MiniGPT-4. Nevertheless, our approach still falls short of the performance demonstrated by LLaVA, which fine-tuned the LLM end-to-end. As an avenue for future research, enhancing the model's capability to process unimodal input with efficacy could be considered.

\begin{figure}[h]
  \centering
  % \vspace{-0.2cm}
     \includegraphics[width=\linewidth]{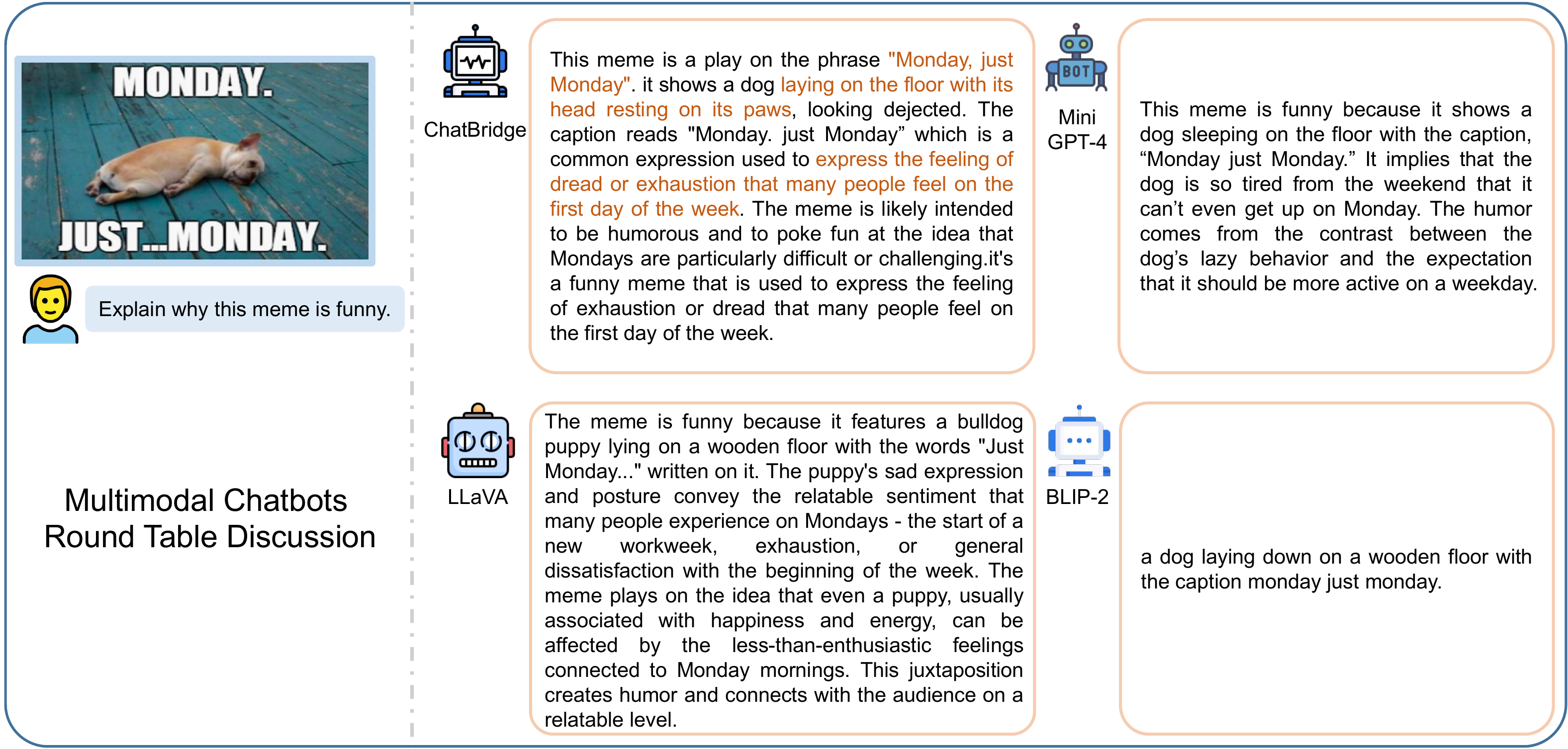}
     % \vspace{-0.5cm}
  \caption{A comparison of SoTA models' responses to an image-based reasoning task.}
  % \vspace{-0.5cm}
\label{fig:chatimage}
\end{figure}

\paragraph{Chat with Multimodal Input}
We also examine the ability of ChatBridge to perform human-assistant chat on multimodal inputs. 
Firstly, we construct a high-quality test set of multimodal chat data by following a similar pipeline as our MULTIS. Specifically, we employ GPT-4~\cite{gpt4} to produce a pool of candidate samples, where we manually choose 90 high-quality samples based on the following criteria: diversity of instructions, correctness of responses, and reliance on different modalities. The selected samples are categorized into three types: reasoning (solving problems through logical thinking and analysis), understanding (comprehending information and interpreting its meaning), and knowledge-based (applying specific knowledge to answer questions accurately).

To illustrate the effectiveness of our model in processing multimodal inputs, we primarily compare the performance of our model when exposed to different modality inputs. 
Similar to the above evaluation, we use GPT-4 to score the responses of each method. According to the evaluation results reported in Table~\ref{table-multichat}, the model incorporating both audio and video inputs outperforms the model relying solely on unimodal inputs on all three sub-tasks. This finding underscores the model's ability to effectively harness the complementary information inherent in each modality, thereby facilitating a more comprehensive comprehension of the data.

\paragraph{Qualitative Analysis}

\begin{figure}[h]
  \centering
     \includegraphics[width=\linewidth]{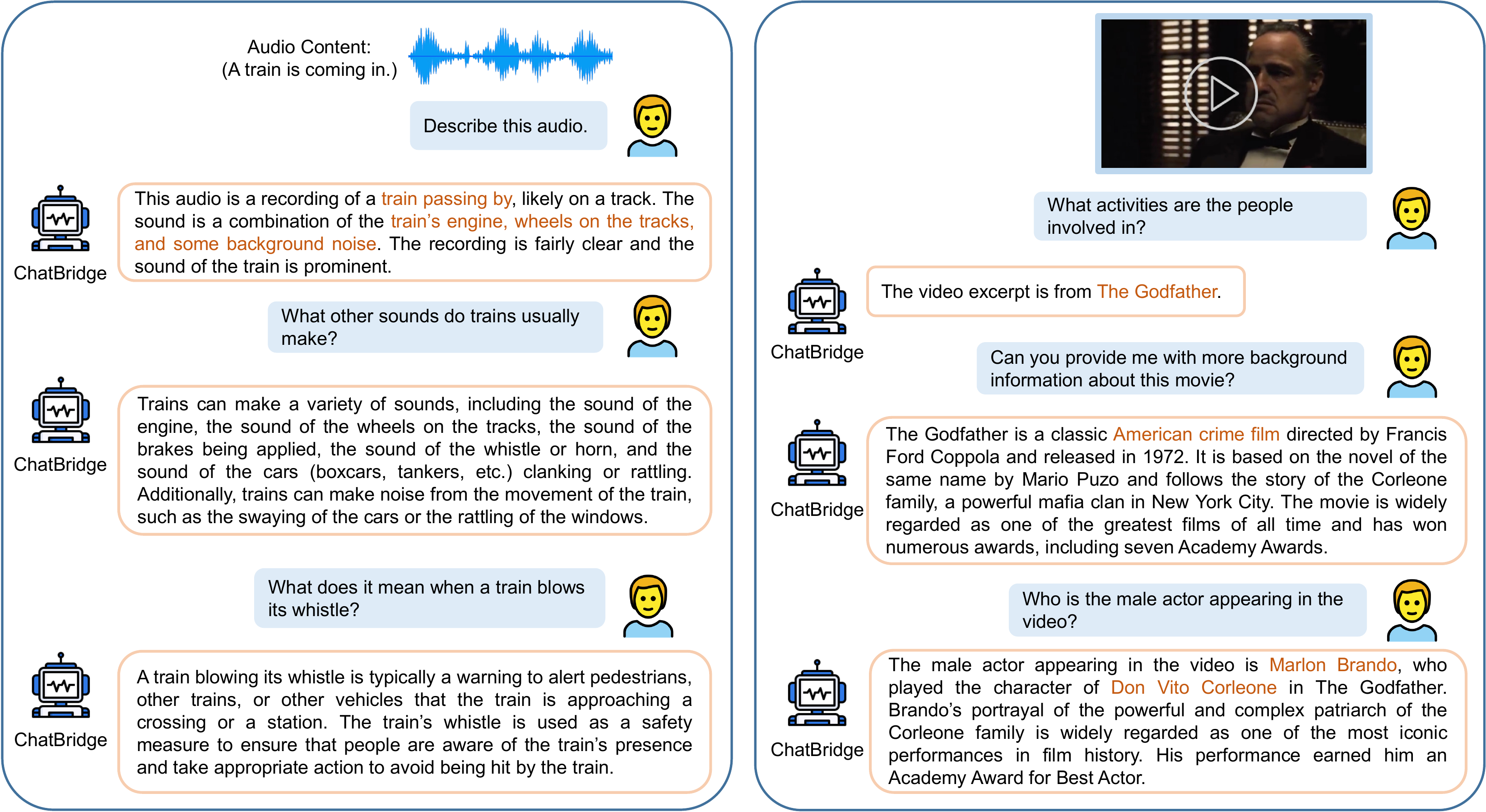}
    \vspace{-0.4cm}
  \caption{Multi-round conversation cases with audio or video inputs.}
  % \vspace{-0.5cm}
\label{fig:chat_audio_video}
\end{figure}

\begin{figure}[h]
    \centering
    \begin{minipage}[t]{0.48\textwidth}
    \centering
    \includegraphics[width=\textwidth]{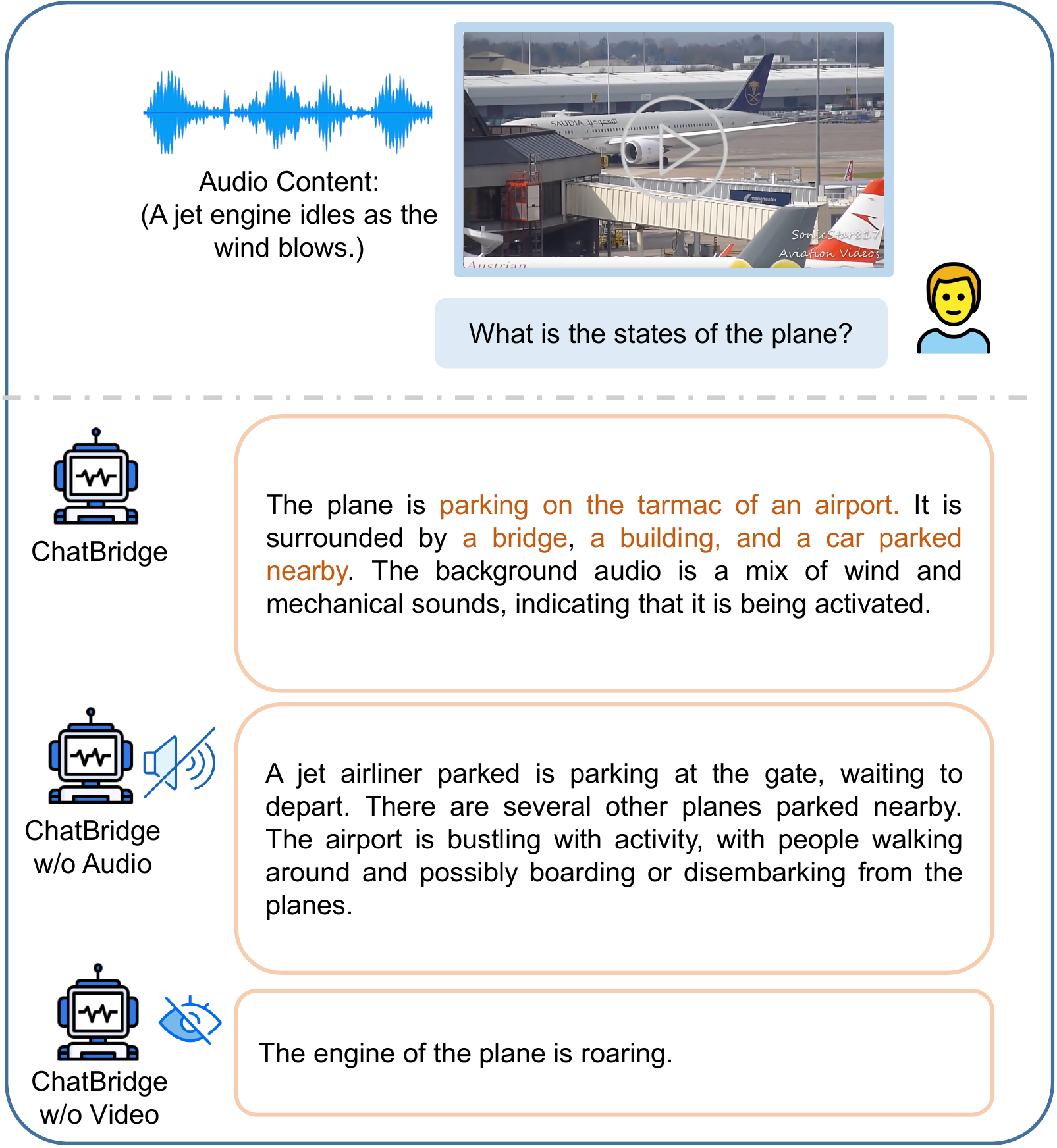}
      \label{fig:chatav_comp}
      \vspace{-0.5cm}
  \caption{A comparison of ChatBridge's responses to different input modalities  with a video-audio input.}
    \end{minipage}\hspace{.15in}
    \begin{minipage}[t]{0.48\textwidth}
    \centering
    \includegraphics[width=\textwidth]{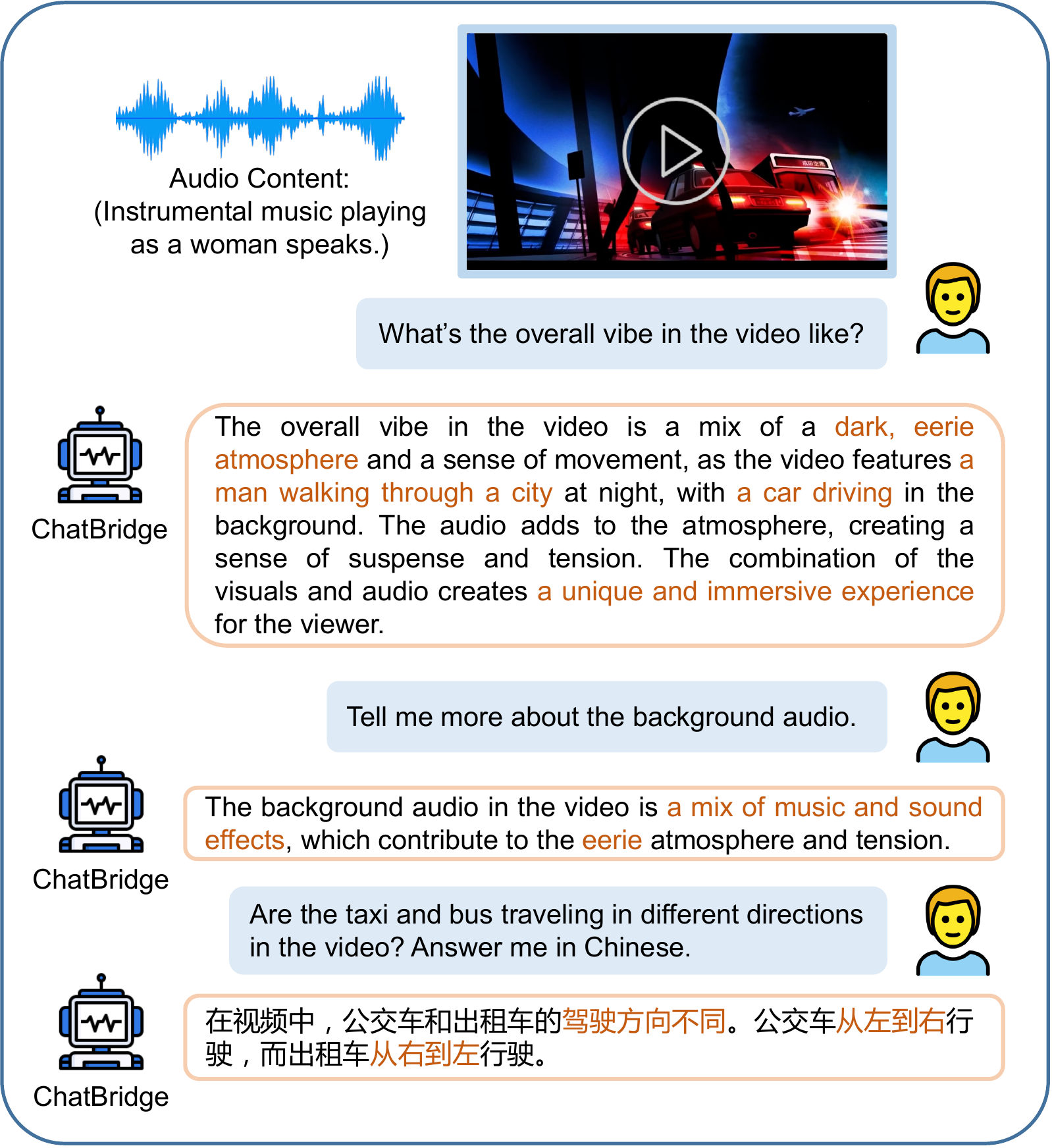}
    \vspace{-0.5cm}
  \caption{A multi-round conversation case with a video-audio input.}
  \label{fig:chatav_multiround}
    \end{minipage}
    \vspace{-0.5cm}
\end{figure}

Furthermore, Figure~\ref{fig:chatimage} depicts ChatBridge's ability to generate more comprehensive responses when both video and audio modalities are present, providing evidence for its competence in co-relating and cooperating different modalities. Finally, the multi-round conversation cases illustrated in Figure~\ref{fig:chat_audio_video} and Figure~\ref{fig:chatav_multiround} demonstrate ChatBridge's efficacy in following human instructions and comprehending both unimodal and multimodal contents.

In Figure~\ref{fig:chatimage}, we show an image-base reasoning case to SoTA image-based LLMs, and compare their responses. Our ChatBridge accurately recognizes the text in the image and almost perfectly understands the humor of the image. 
Furthermore, Figure\ref{fig:chatimage} depicts ChatBridge's ability to generate more comprehensive responses when both video and audio modalities are present, providing evidence for its competence in co-relating and cooperating different modalities.
Finally, the multi-round conversation cases illustrated in Figure~\ref{fig:chat_audio_video} and Figure~\ref{fig:chatav_multiround} demonstrate ChatBridge's efficacy in following human instructions and comprehending both unimodal and multimodal contents.

\section{Discussions}
This paper proposes ChatBridge, a multimodal language model capable of interpreting, correlating, and reasoning about various modalities through two-modality data paired with language. We introduce MULTIS, a multimodal instruction tuning dataset, to instruction-tune our model, which coveres a wide range of multimodal tasks in text, image, video, and audio modalities. Consequently, our model exhibits strong and noteworthy quantitative results on zero-shot multimodal tasks. Moreover, qualitative examples demonstrate ChatBridge’s diverse capabilities in following multimodal instructions, such as complex multimodal reasoning, knowledge-grounded multimodal understanding, and multi-turn conversations. 
These contributions and findings may pave the way for future research into building general-purpose multimodal assistants.

\paragraph{Limitations} Here, we describe some limitations of our model as well as opportunities for further improving our model. Specifically, we note the following: 1) We observe that our model exhibits weaknesses in understanding and grounding long-range videos and audios, necessitating a more precise temporal modeling approach. 2) Our framework can be extended to incorporate additional modalities, such as sketch and point cloud. 
3) Though the frozen modules in our framework alleviate computational burden, they may lead to insufficient performance and also introduce prior biases from pretrained models. 

% \bibliographystyle{plain}
% \bibliography{sample}

\clearpage

\appendix

\section{More Details about MULTIS Dataset}
\begin{table}[h]
\vspace{10mm}
\renewcommand\arraystretch{3}

    \centering

 \resizebox{\linewidth}{!}{

\begin{tabular}{lllll}
\toprule
Modality                     & Dataset                     & Type           & Held-Out           & Description \\ \midrule
\multirow{7}{*}{Image}       & COCO Caption~\cite{coco}   & Caption  &   \ding{55}    &    \makecell[l]{We use Karpathy~\cite{karpathy} split, which divides caption data into 82k/5k/5k images for train/val/test sets, \\ with each image corresponding to 5 captions.}      \\\cmidrule{2-5} 
                             &    Flickr30k Caption~\cite{flickr}              & Caption       &  \ding{51} (test) & 
\makecell[l]{We use test split in Flickr30k dataset, which contain 1k images,  with each image corresponding to\\ 5 captions.}     \\ \cmidrule{2-5} 
                             &      NoCaps~\cite{agrawal2019nocaps}     & Caption       &      \ding{51} (val)   &  
\makecell[l]{NoCaps dataset consists of 166,100 human-generated captions describing 15,100 images from the\\ Open Images~\cite{openimage} validation and test sets. We use the val set in NoCaps dataset.}     \\ \cmidrule{2-5} 
                             & VQAv2~\cite{vqa}   & QA             &  \ding{55}   &          
\makecell[l]{VQAv2 dataset contains 265,016 images,  with 5.4 questions on average per image. Each question \\has 10 ground truth answers.}     \\ \cmidrule{2-5} 
                             &   QA~\cite{vg}      & QA  & \ding{55}    &        
\makecell[l]{Visual Genome consists of 101,174 images from MSCOCO~\cite{coco} with 1.7 million QA pairs, with \\17 questions per image on average. }     \\ \cmidrule{2-5} 
                             &    GQA~\cite{gqa}      & QA  & \ding{51} (test-dev) &           
\makecell[l]{GQA dataset is a visual question answering dataset with real images from the Visual Genome~\cite{vg} \\dataset and balanced question-answer pairs. We use the balanced test-dev split in GQA dataset.}     \\ \cmidrule{2-5} 
                             &  OKVQA~\cite{okvqa}   & QA  & \ding{51} (test) &             
\makecell[l]{OK-VQA is a dataset for visual question answering that requires methods that can draw upon \\outside knowledge to answer questions. OK-VQA dataset contains 14,055 open-ended questions \\and with 5 ground truth answers per question. We use the test split in OK-VQA dataset.}           \\ \midrule
\multirow{4}{*}{Video}       & MSRVTT Caption~\cite{msrvtt}   & Caption  & \ding{55}  &            
\makecell[l]{MSR-VTT dataset consists of 10,000 video clips from 20 categories, and each clip is annotated \\with 20 sentences. The standard splits use 6,513 clips for training, 497 clips for validation, and\\ 2,990 clips for testing. }     \\ \cmidrule{2-5} 
                             &    VATEX Caption~\cite{vatex} & Caption      &  \ding{51} (test)    &          
\makecell[l]{We use the test split in VATEX dataset, which contains 6000 videos with 10 captions per video.}   \\ \cmidrule{2-5} 
                             & MSRVTT QA~\cite{msrvtt} & QA           &  \ding{55}  &  
\makecell[l]{MSR-VTT QA dataset consists of about 158k video QA pairs. }  \\ \cmidrule{2-5}
                             &  MSVD QA~\cite{msvd}     &QA      &  \ding{51} (test)  &  
\makecell[l]{We use test split in MSVD QA dataset, which consists 13k video QA pairs. }     \\ \midrule
\multirow{2}{*}{Audio}       & AudioCaps~\cite{audiocaps}    & Caption           &  \ding{55} &           
\makecell[l]{AudioCaps dataset consists of about 46K audio clips to human-written text pairs collected via crowd-\\ -sourcing on the AudioSet~\cite{audioset} dataset. }     \\ \cmidrule{2-5} 
                             &    Clothov2~\cite{drossos2020clotho}       & Caption   &   \ding{51} (test)     &            
\makecell[l]{Clotho consists of 6974 audio samples, and each audio sample has 5 captions. We use the val split\\ in Clotho datasets. }     \\ \midrule
\multirow{3}{*}{Video-Audio} & VALOR32K~\cite{valor}        & Caption  & \ding{51}(test)   &          
\makecell[l]{VALOR-32K is an audio-visual captioning dataset. In this dataset, each video corresponds to a caption \\consisting of both audio and visual contents. VALOR-32K is split into 25K/3.5K/3.5K videos for \\training, validation, and testing. We use the test split for evaluation.}     \\ \cmidrule{2-5} 
                             & MUSIC AVQA~\cite{musicqa}    & QA    & \ding{51} (test) &      
\makecell[l]{MUSIC AVQA is an audio-visual question answering (AVQA) dataset, which aims to answer ques-\\-tions regarding different visual objects, sounds, and their associations in videos. It contains 9.3K\\ videos. We use the test split for evaluation.}     \\ \cmidrule{2-5} 
                             & AVSD~\cite{avsd}               & Dialogue &  \ding{51} (DSTC7 test)     &  
\makecell[l]{AVSD is a audio-visual dataset for dialogue understanding, which aims to generate responses in a \\dialog about a video, given the dialog history and audio-visual content of the video. We use the test\\ split in DSTC7 Track for evaluation.}     \\ \bottomrule  \\     
\end{tabular}}

\caption{Compositions of task-specific data in our multimodal instruction dataset MULTIS.}

\end{table}

\begin{table}[h]
\centering
\renewcommand\arraystretch{1.5}

 \resizebox{0.9\linewidth}{!}{
\begin{tabular}{lllcl}
\toprule
Modality &
  Source Dataset &
  Type &
  \#Samples &
  Description \\ \midrule
\multirow{3}{*}{Image} &
  \multirow{3}{*}{MSCOCO~\cite{coco}} &
  Detailed Description &
  26k &
  A rich and comprehensive description for an image \\ \cmidrule{3-5} 
 &
   &
  Conversation &
  58k &
  \makecell[l]{Multi-turn conversation between the assistant and a person\\ asking questions about the image} \\ \cmidrule{3-5} 
 &
   &
  Reasoning &
  77k &
  \makecell[l]{In-depth reasoning questions about image contents which\\ requires a step-by-step reasoning process to answer} \\ \midrule
\multirow{3}{*}{Video} &
  \multirow{3}{*}{MSRVTT~\cite{msrvtt}} &
  Detailed Description &
  10k &
 \makecell[l]{ A rich and comprehensive description for a video, including\\ more motional information} \\  \cmidrule{3-5} 
 &
   &
  Conversation &
  10k &
  \makecell[l]{Multi-turn conversation between the assistant and a person\\ asking questions about the video} \\ \cmidrule{3-5}
 &
   &
  Reasoning &
  10k &
  \makecell[l]{In-depth reasoning questions about video contents which \\requires a step-by-step reasoning process to answer} \\ \midrule
Audio &
  AudioCaps~\cite{audiocaps} &
  Conversation &
  9k &
  \makecell[l]{Multi-turn conversation between the assistant and a person\\ asking questions about the audio} \\ \midrule
\multirow{2}{*}{Video-Audio} &
  \multirow{2}{*}{VALOR~\cite{valor}} &
  Detailed Description &
  9k &
  \makecell[l]{A rich and comprehensive description for a video together\\ with its background audio} \\ \cmidrule{3-5}
 &
   &
  Conversation &
  9k &
  \makecell[l]{Multi-turn conversation between the assistant and a person \\asking questions about the video and its background audio} \\ \bottomrule \\
\end{tabular}}
\caption{Compositions of multimodal chat data in MULTIS. We use the image instruction dataset provided by LLaVA~\cite{llava} as the chat data for image modality.}
\end{table}

\begin{table}[h]
\centering
\renewcommand\arraystretch{2}
 \resizebox{\linewidth}{!}{
\begin{tabular}{lll}
\toprule
Modality                     & Type     & Examples of Instruction Templates   \\ \midrule
\multirow{2}{*}{Image}       & Caption  & 
\makecell[l]{
Generate a brief sentence to describe the content of the image. \\
Write a single sentence that conveys what the image depicts. \\
Summarize the image content in a single sentence.}             \\ \cmidrule{2-3} 
                             & QA  & 
 \makecell[l]{
Use the visual aid to respond to the question briefly: \textit{<QUESTION>}\\
Analyze the picture and provide a brief answer to \textit{<QUESTION>} \\
Use the information presented in the image to shortly answer \textit{<QUESTION>}}            \\ \midrule
\multirow{2}{*}{Video}       & Caption  & 
 \makecell[l]{
Generate a concise description for this video.\\
Give a brief overview of the information presented in this video. \\
Write a short summary that effectively conveys the main message of this video.}     \\ \cmidrule{2-3} 
                             & QA  & 
 \makecell[l]{
Based on the information presented in the video, provide a short answer to question \textit{<QUESTION>}\\
Analyze the video and provide a one-word answer to question \textit{<QUESTION>} \\
With the aid of the given video, what is your simple answer to \textit{<QUESTION>}}    \\ \midrule
Audio                        & Caption  & 
 \makecell[l]{
Listen to this audio and summarize its content in one sentence.\\
Write a succinct summary of the key takeaways from this audio. \\
After listening to the audio, generate a one-sentence overview of its main ideas.}  \\ \midrule
\multirow{3}{*}{Video-Audio} & Caption  & 
\makecell[l]{
Combining the audiovisual information of this video, generate a sentence to describe its content.\\
Synthesize the audio and visual data in this video to create a sentence that encapsulates its meaning. \\
Describe the content of this video by integrating the audio and visual elements into a single sentence.}     \\ \cmidrule{2-3} 
                             & QA  & 
\makecell[l]{ Based on the video and audio, could you provide a short answer to question: \textit{<QUESTION>} \\
Utilizing the video and audio content, briefly respond to \textit{<QUESTION>}\\
Analyze the video and audio and give me a short answer about \textit{<QUESTION>}
}  \\ \cmidrule{2-3} 
                             & Dialogue & 
\makecell[l]{
Based on the video and audio, answer my following questions.\\
Synthesize the audio and visual data in this video and answer my questions.\\
Given this video together with its background audio, answer my next questions.
} \\ \bottomrule \\
\end{tabular}}
\caption{Examples of instruction templates for constructing task-specific data in MULTIS. We show 3 examples for each task.
We use prompts: "Give image: \textit{<image input>}.", "Give audio: \textit{<audio input>}.", "Give video: \textit{<video input>}.", and "Give video: \textit{<video input>} and its background audio: \textit{<audio input>}." to deal with different modality inputs.}
\end{table}

\clearpage
\section{Example Samples of MULTIS Dataset}

\begin{table}[h]
\centering
  \begin{minipage}{\textwidth}
\centering  
\vspace{10mm}
\scalebox{0.88}{
\begin{tabular}{l p{12.5cm} }
\toprule
 \multicolumn{2}{l}{\bf Audio Conversation Sample1:}  \\ \midrule
 \multicolumn{2}{c}{\includegraphics[width=\linewidth]{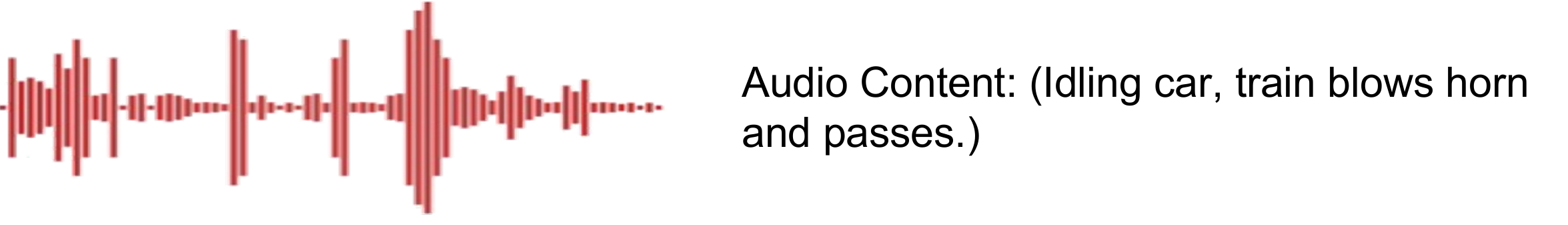} }
\\ \midrule
Human & Where is the audio recording location?
 \\
Assistant & It should be near the train station, because \textcolor{red}{I hear the sound of the train whistle.}
\\ \midrule
Human & Do you hear any other voices?
 \\
Assistant & Yes, I hear \textcolor{red}{the car engine idles.}
\\ \bottomrule
\\
 \\ \toprule
\multicolumn{2}{l}{\bf Audio Conversation Sample2:}  \\ \midrule
 \multicolumn{2}{c}{\includegraphics[width=\linewidth]{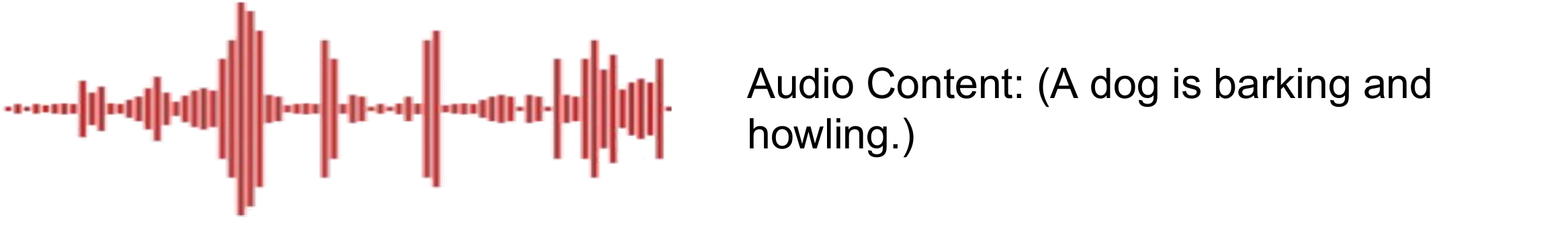} }
\\ \midrule
Human & Can you hear the sounds of animals?
 \\
Assistant & Yes, I heard a \textcolor{red}{dog's voice.}
\\ \midrule
Human & Do you have further detail about it?
 \\
Assistant & This dog seems very angry because it is \textcolor{red}{barking and hawking.}
\\

\bottomrule \\
\end{tabular}
}
  \caption{Examples of audio-input samples in MULTIS. We use \textcolor{red}{red} color to highlight audio-related content in generated data.} 
 
  \end{minipage}

\end{table}

\begin{table}[h]
  \begin{minipage}{0.99\textwidth}
\centering  
\vspace{-4mm}
\scalebox{0.88}{
\begin{tabular}{l p{12.5cm} }
\toprule
 \multicolumn{2}{l}{\bf Video Conversation
 Sample:}  \\
\midrule
 \multicolumn{2}{c}{\includegraphics[width=0.8\linewidth]{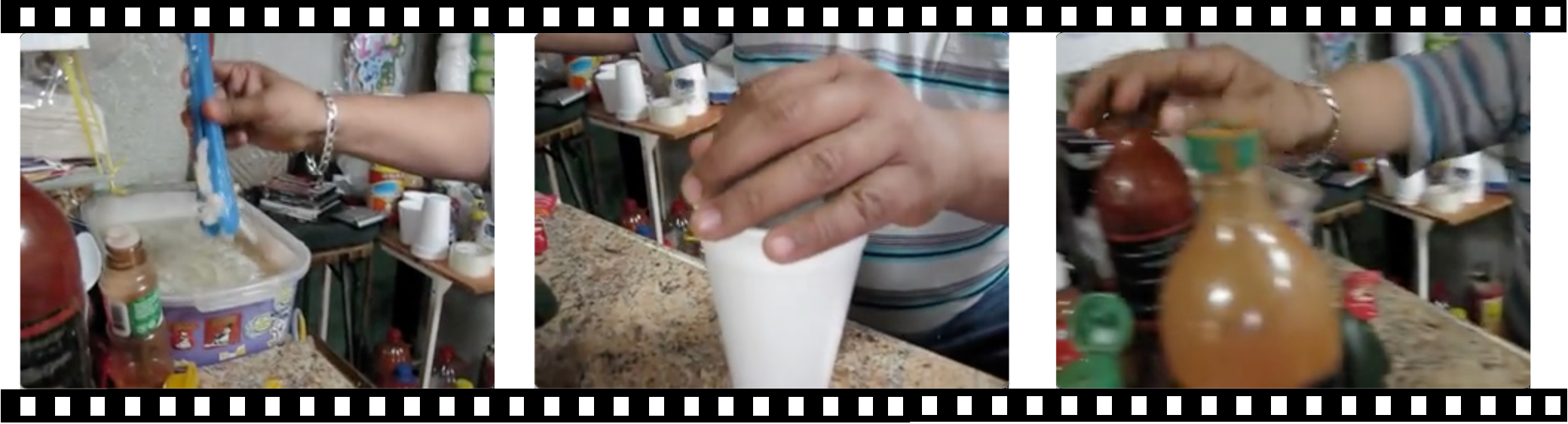} }
\\ \midrule
Human & What is the primary action happening in the video?
 \\
Assistant & In this video, a man is \textcolor{green1}{making a cup of noodles in the kitchen.}
\\ \midrule
Human & What kind of cup is he using?
 \\
Assistant & He's using \textcolor{green1}{a styrofoam cup} to make the noodles.
\\ \midrule
Human & Is the man adding any other ingredients to the cup?
 \\
Assistant & Yes, the man is also \textcolor{green1}{adding onions} to the cup of noodles.
\\ \midrule
Human & What is the likely purpose of this video?
 \\
Assistant & The purpose of this video is to \textcolor{green1}{demonstrate a simple, quick meal preparation process} using a styrofoam cup and noodles, possibly to serve as a tutorial for others who would like to give it a try.
\\
\bottomrule
 \\ 
\\
\toprule
 \multicolumn{2}{l}{\bf Video Detailed Description
 Sample:}  \\ \midrule
 \multicolumn{2}{c}{\includegraphics[width=0.8\linewidth]{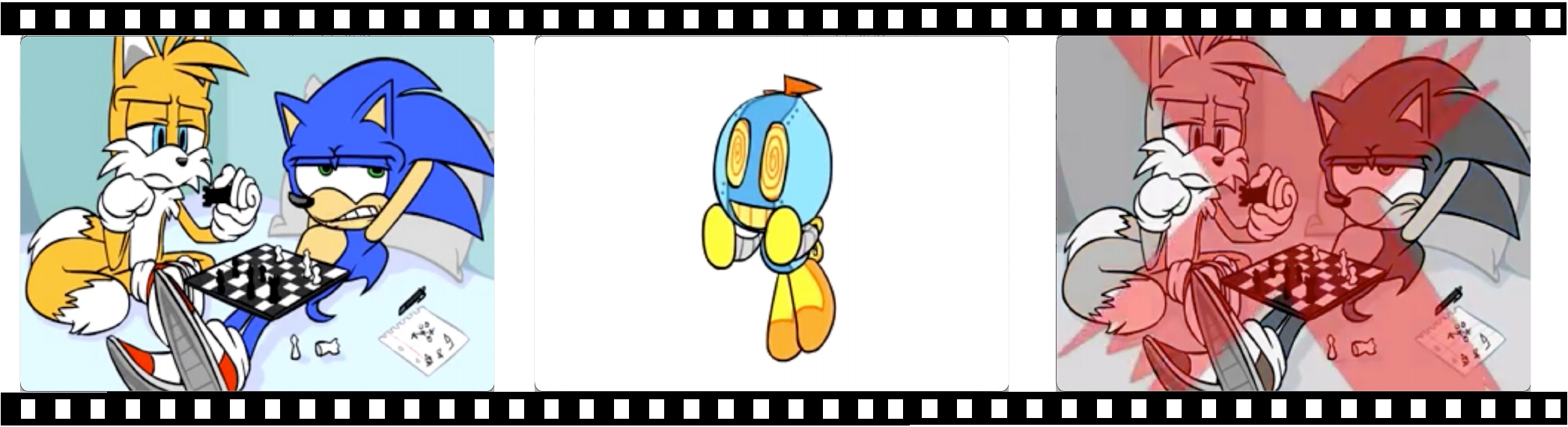} }
\\ \midrule
Human & What are the key elements in this video?
 \\
Assistant & The video showcases \textcolor{green1}{various animated characters and scenes} in different settings. It starts with \textcolor{green1}{Sonic and Tails playing a game of chess} while surrounded by \textcolor{green1}{a graffiti-covered wall}. Then, \textcolor{green1}{a scene shifts to another cartoon character wearing a blue and yellow hat, flying a kite with a smiling face, and an orange flag atop a pole.} The video transitions to \textcolor{green1}{a black and white scene featuring a red cross}. Sonic the Hedgehog is featured in the video too while advertising a new game.
\\
\bottomrule
 \\ 
\\
\toprule
 \multicolumn{2}{l}{\bf Video Reasoning
 Sample:}  \\ \midrule
 \multicolumn{2}{c}{\includegraphics[width=0.8\linewidth]{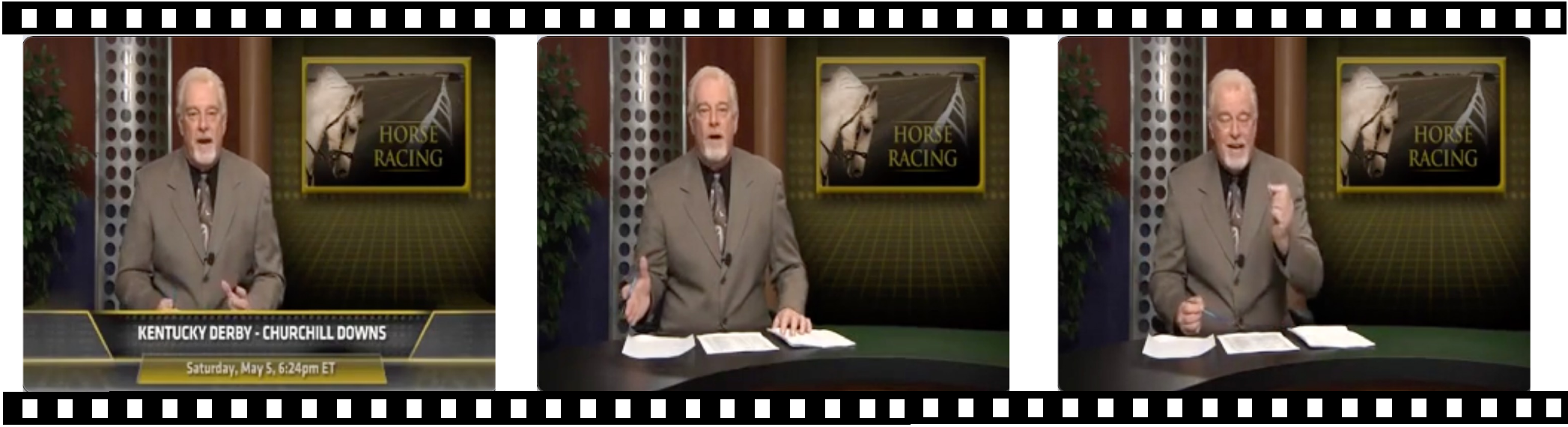} }
\\ \midrule
Human & What can be inferred about the man's profession and the purpose of his discussion?
 \\
Assistant & Based on the video, \textcolor{green1}{the man is dressed in a suit and tie sitting at a table surrounded by papers}. He is talking about the \textcolor{green1}{Kentucky Derby and horse racing.} It can be inferred that he is a professional sports analyst or an expert in horse racing. The purpose of his discussion is to inform the viewers about the race, share his insights on the competition, handicap potential outcomes, and provide interesting background information on the event.
\\
\bottomrule \\
\end{tabular}
}

\caption{Examples of video-input samples in MULTIS. We use \textcolor{green1}{green} color to highlight video-related content in generated data.}  

  \end{minipage}
\end{table}

\begin{table}[h]
  \begin{minipage}{0.99\textwidth}
\centering  
\vspace{-4mm}
\scalebox{0.88}{
\begin{tabular}{l p{12.5cm} }
\toprule

\multicolumn{2}{l}{\bf Video-Audio Conversation
 Sample1:}  \\
\midrule
 \multicolumn{2}{l}{\includegraphics[width=\linewidth]{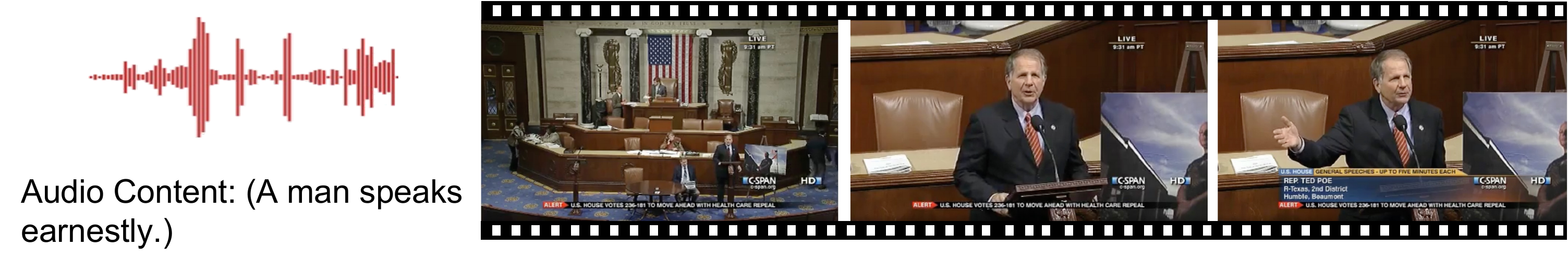} }
\\ \midrule
Human & What is the man at the podium doing?
 \\
Assistant & The man at the podium is \textcolor{red}{speaking earnestly}, probably \textcolor{red}{giving a speech or addressing an audience}.
\\
\midrule
Human & Please describe the man's outfit.
 \\
Assistant &  The man is \textcolor{green1}{wearing a suit and tie, and the tie is red and grey striped}.
\\ \midrule
Human & What else can be seen in the room?
 \\
Assistant & In the room, there is a man standing, \textcolor{green1}{a brown seat back, a table with a stack of papers and a leather chair.}
\\ \bottomrule
 \\ 
 \\
\toprule
 \multicolumn{2}{l}{\bf Video-Audio Conversation
 Sample2:}  \\ \midrule
 \multicolumn{2}{l}{\includegraphics[width=\linewidth]{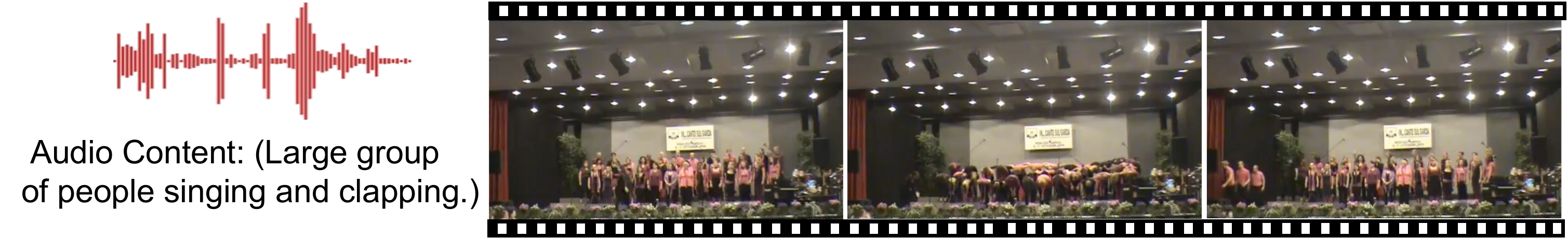} }
\\ \midrule
Human & What type of event is taking place in the video?
 \\
Assistant & \textcolor{green1}{A choir is performing} at a church event.
\\
\midrule
Human & Can you describe some of the people in the choir?
 \\
Assistant &  Certainly! Some members of the choir include a person \textcolor{green1}{wearing a pink shirt}, a woman wearing \textcolor{green1}{a pink jacket}, a person in \textcolor{green1}{a plaid shirt}, and a person \textcolor{green1}{in a red shirt}.
\\ \midrule
Human & What is the audience's reaction to the choir's performance?
 \\
Assistant & The audience sounds very appreciative, as \textcolor{red}{I can hear a large group of people clapping.}
\\ \bottomrule
 \\  \\

\toprule
 \multicolumn{2}{l}{\bf Video-Audio Detailed Description
 Sample:}  \\ \midrule
 \multicolumn{2}{l}{\includegraphics[width=\linewidth]{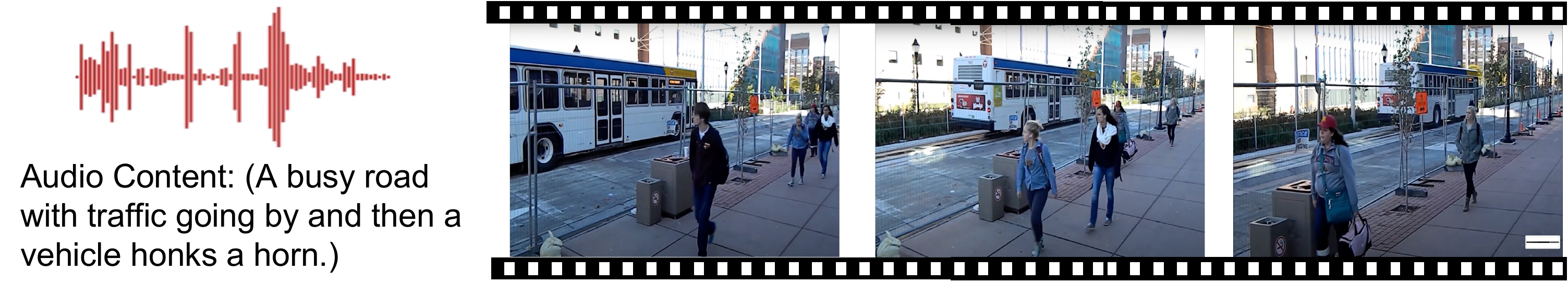} }
\\ \midrule
Human & Write a detailed description of the given video.
 \\
Assistant & In this video, \textcolor{green1}{a busy city street} is filled with \textcolor{red}{the sounds of traffic and a honking horn.} \textcolor{green1}{People are walking on the sidewalk} with some wearing jackets. \textcolor{green1}{A white and blue bus} passes by, and there are various objects along the sidewalk, such as \textcolor{green1}{a gray trash can and a concrete planter.} The street is surrounded by buildings.
\\
\bottomrule \\
\end{tabular}}

\caption{Examples of video-audio input samples in MULTIS. We use \textcolor{green1}{green} and \textcolor{red}{red} colors to highlight video- and audio-related content in generated data.}  

  \end{minipage}
\end{table}

\clearpage
\section{More Case Studies of ChatBridge}
\begin{figure}[htbp]
\centering
\begin{minipage}[t]{0.48\textwidth}
\centering
\includegraphics[width=\textwidth]{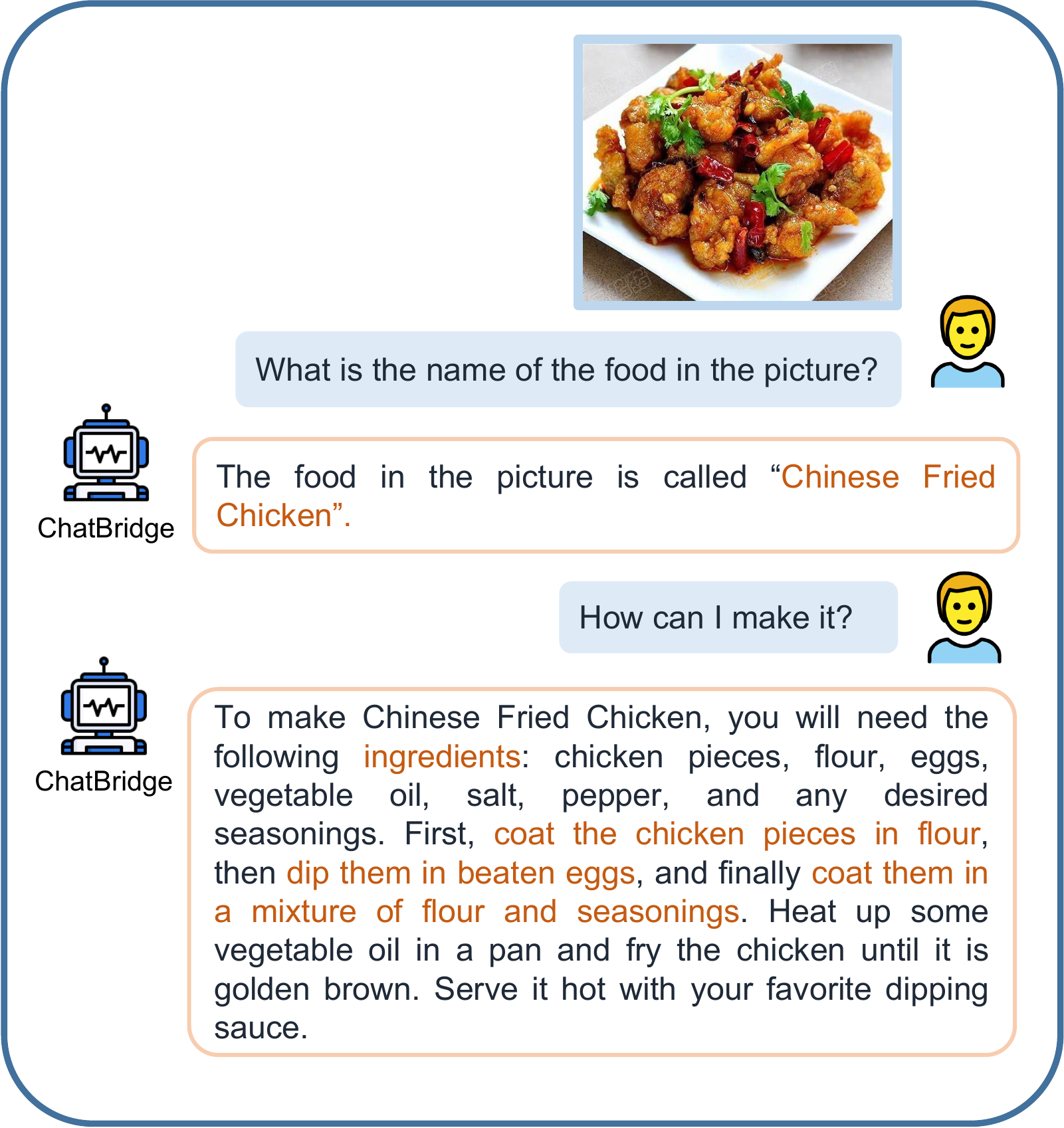}
\caption{Providing advice and step-by-step instructions about an image.}
\end{minipage}\hspace{.15in}
\begin{minipage}[t]{0.48\textwidth}
\centering
\includegraphics[width=\textwidth]{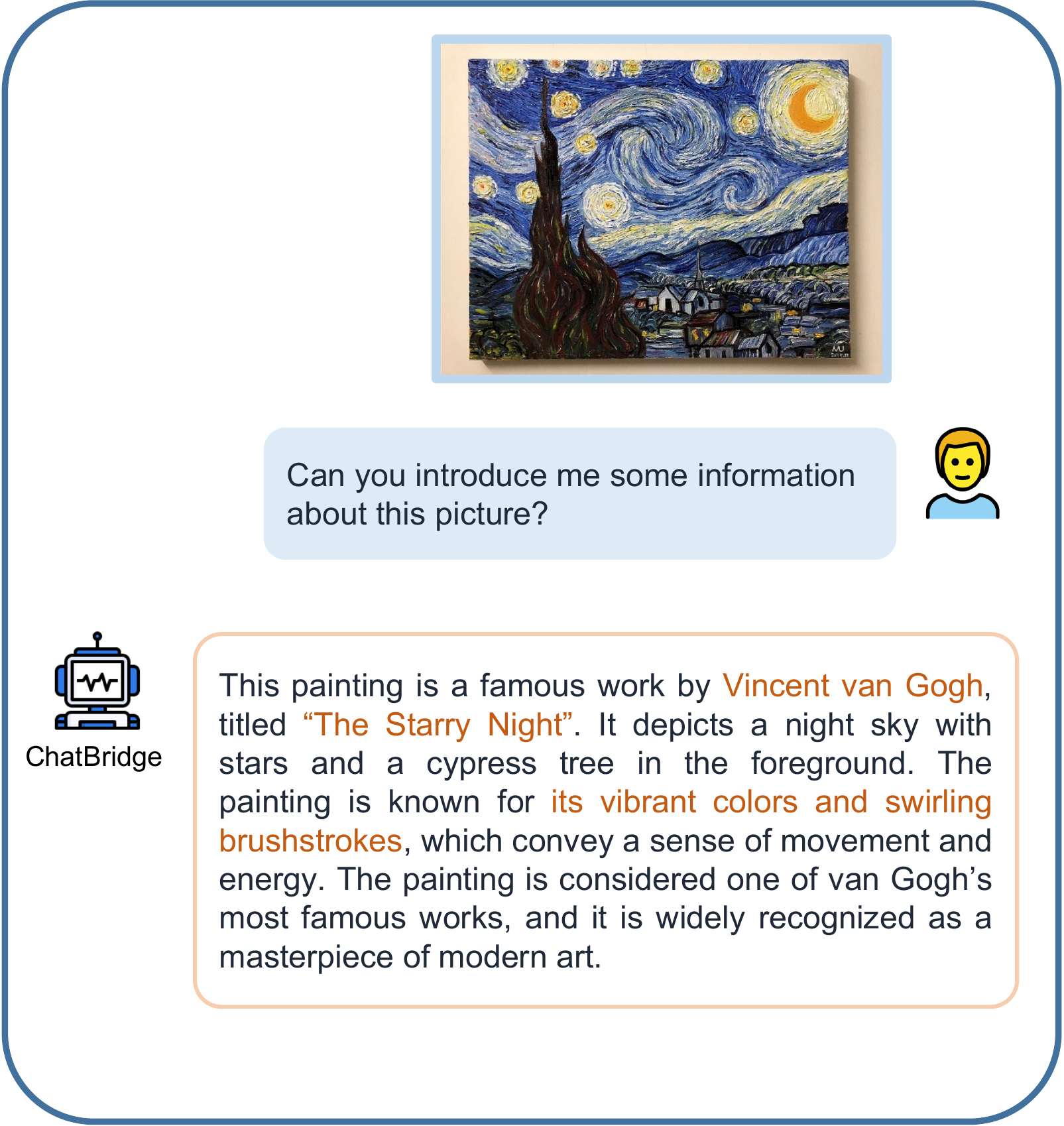}
\caption{Introducing the background knowledge about an image.}
\end{minipage}
\end{figure}

\begin{figure}[htbp]
\centering
\begin{minipage}[t]{0.48\textwidth}
\centering
\includegraphics[width=\textwidth]{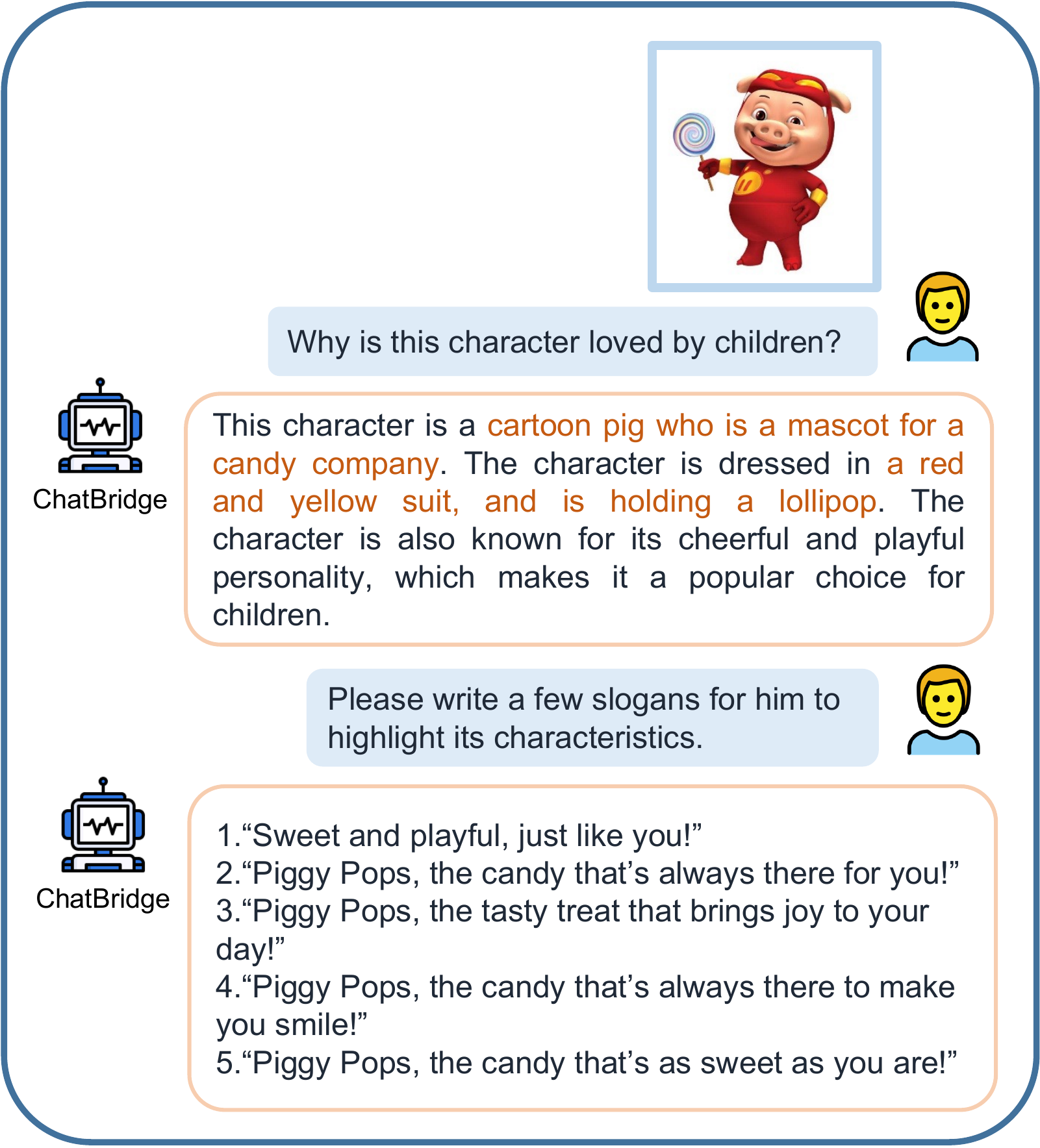}
\caption{Reasoning and creating with an image.}
\end{minipage}\hspace{.15in}
\begin{minipage}[t]{0.48\textwidth}
\centering
\includegraphics[width=\textwidth]{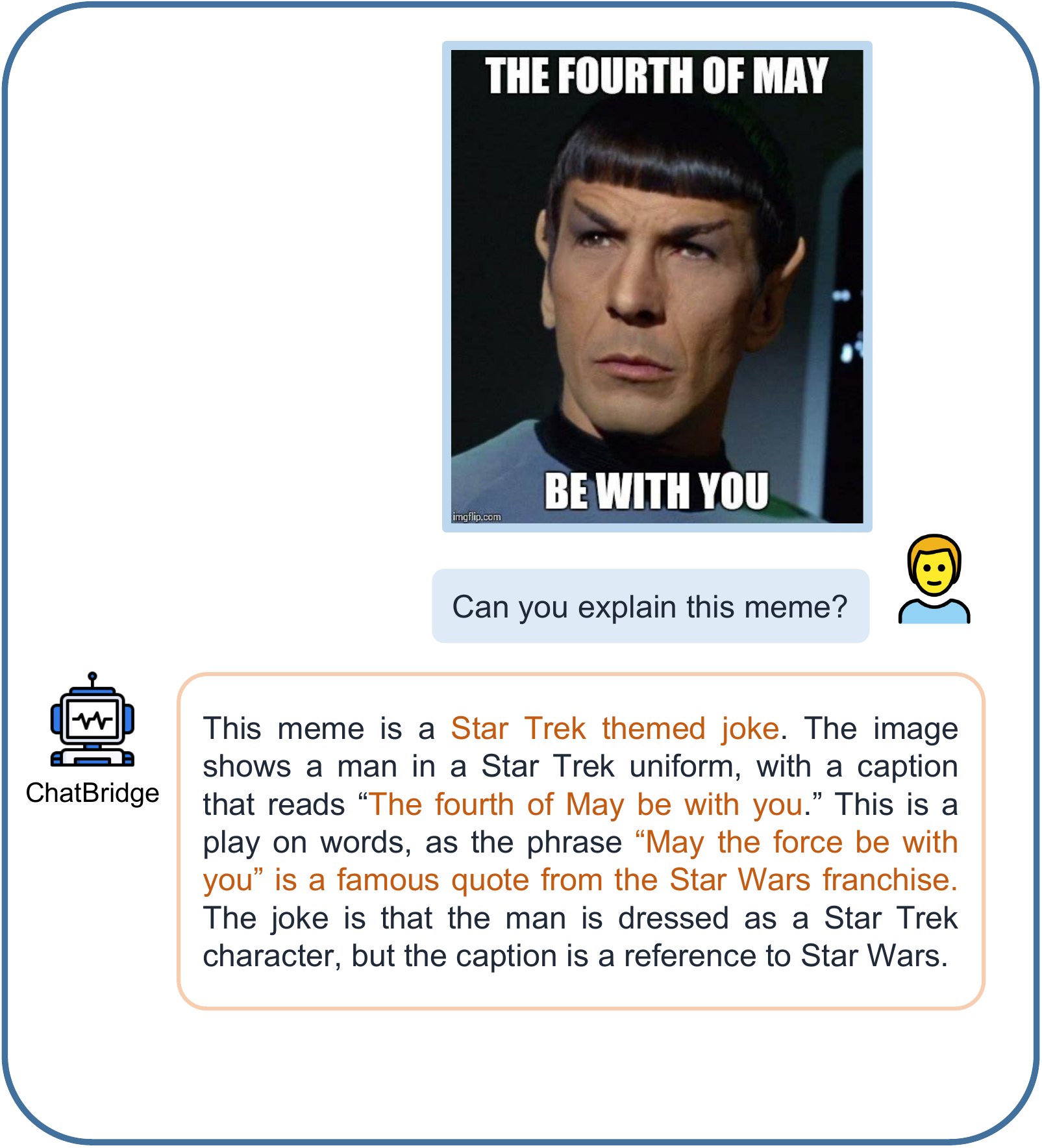}
\caption{Explaining the amusing point in an image.}
\end{minipage}
\end{figure}

\begin{figure}[htbp]
\centering

\includegraphics[width=\linewidth]{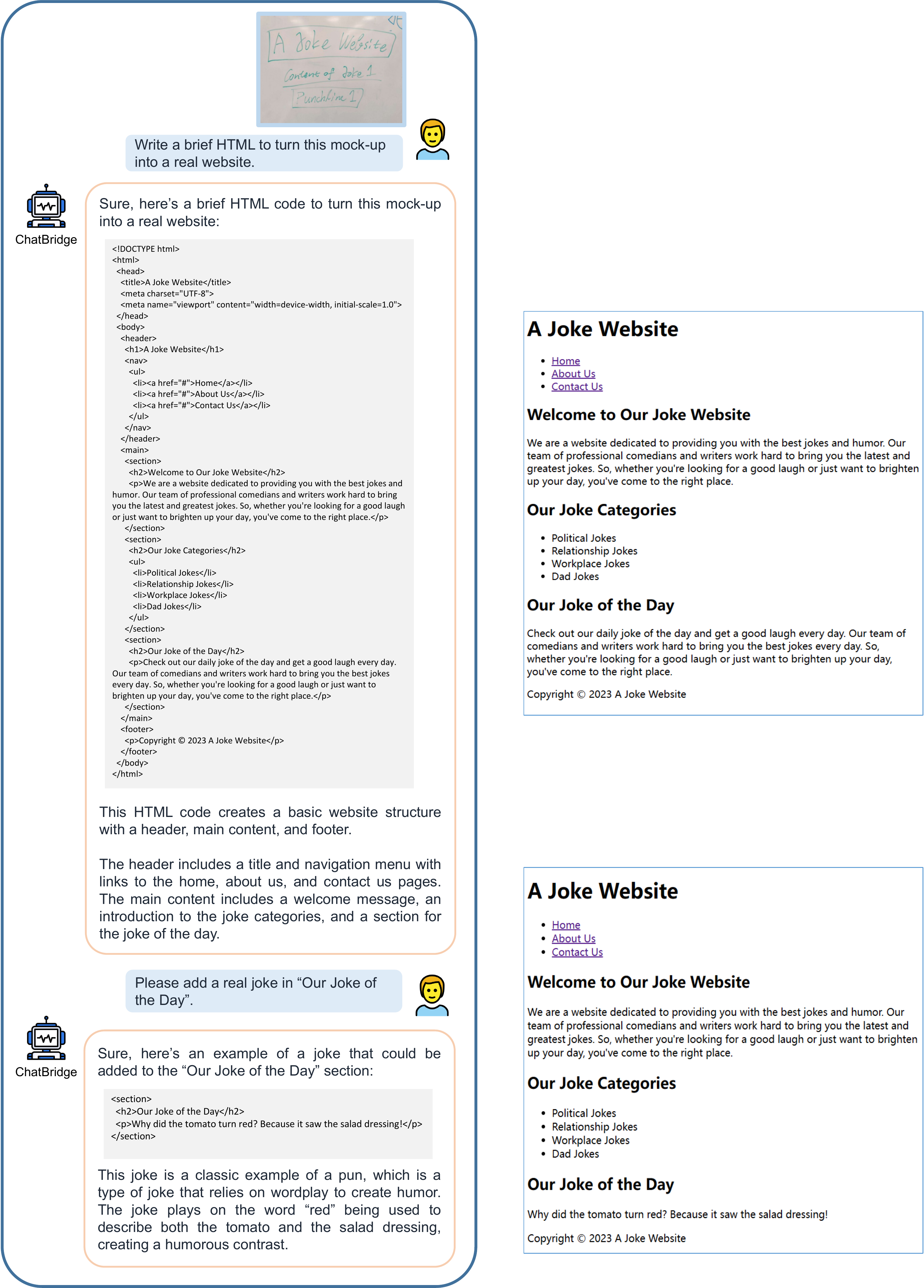}
\caption{Generating HTML code based on a hand-drawn draft and modifying the code according to human opinions. We compile the HTML code in the conversation and show the website in the right column.}

\end{figure}

\begin{figure}[htbp]
\centering
\begin{minipage}[t]{0.48\textwidth}
\centering
\includegraphics[width=\textwidth]{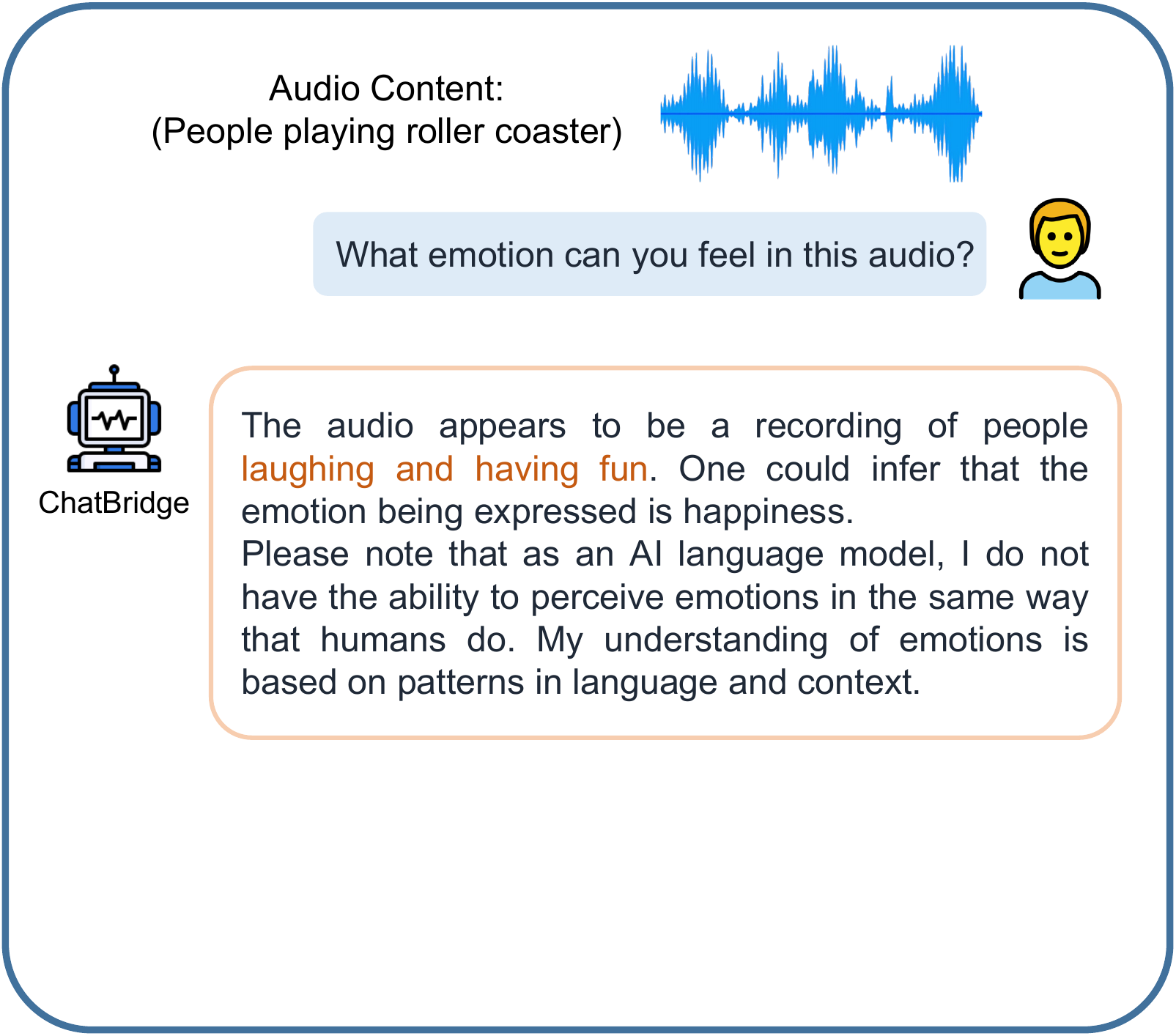}
\caption{Inferring the emotions behind an audio.}
\end{minipage}\hspace{.15in}
\begin{minipage}[t]{0.48\textwidth}
\centering
\includegraphics[width=\textwidth]{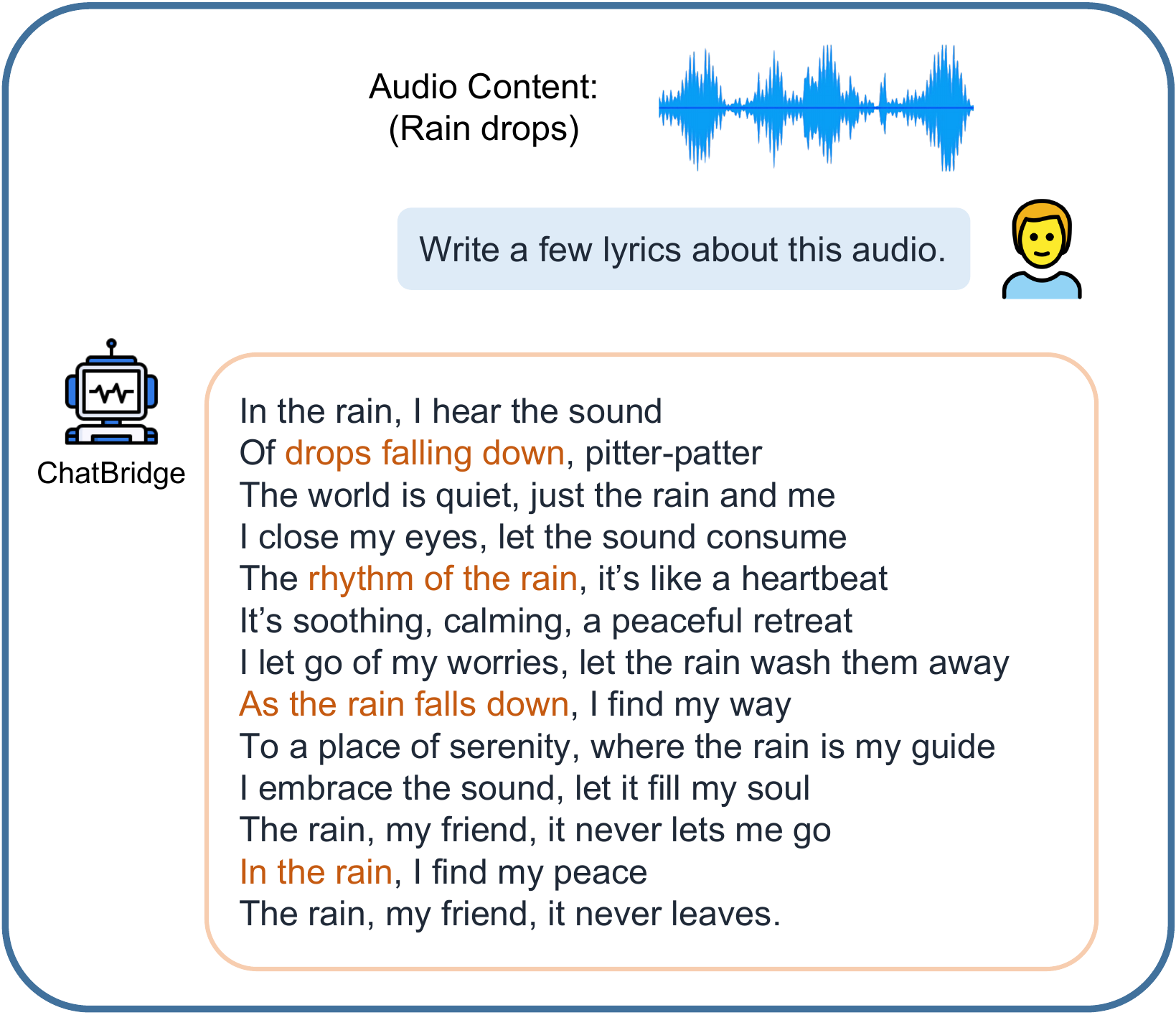}
\caption{Writing lyrics based on audio.}
\end{minipage}
\end{figure}

\begin{figure}[htbp]
\centering
\begin{minipage}[t]{0.48\textwidth}
\centering
\includegraphics[width=\textwidth]{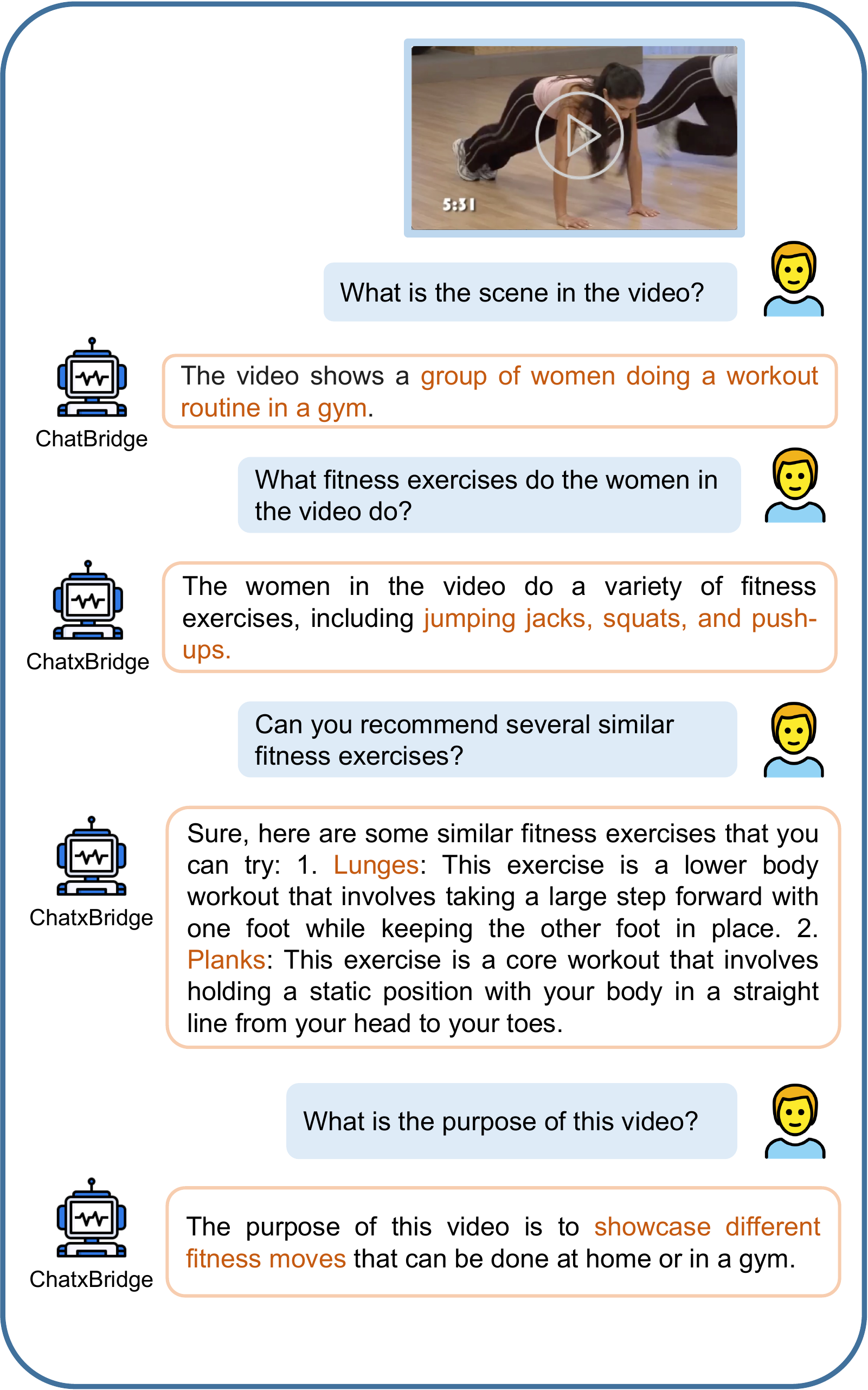}
\caption{Engaging in informative, multi-turn conversations about a video.}
\end{minipage}\hspace{.15in}
\begin{minipage}[t]{0.48\textwidth}
\centering
\includegraphics[width=\textwidth]{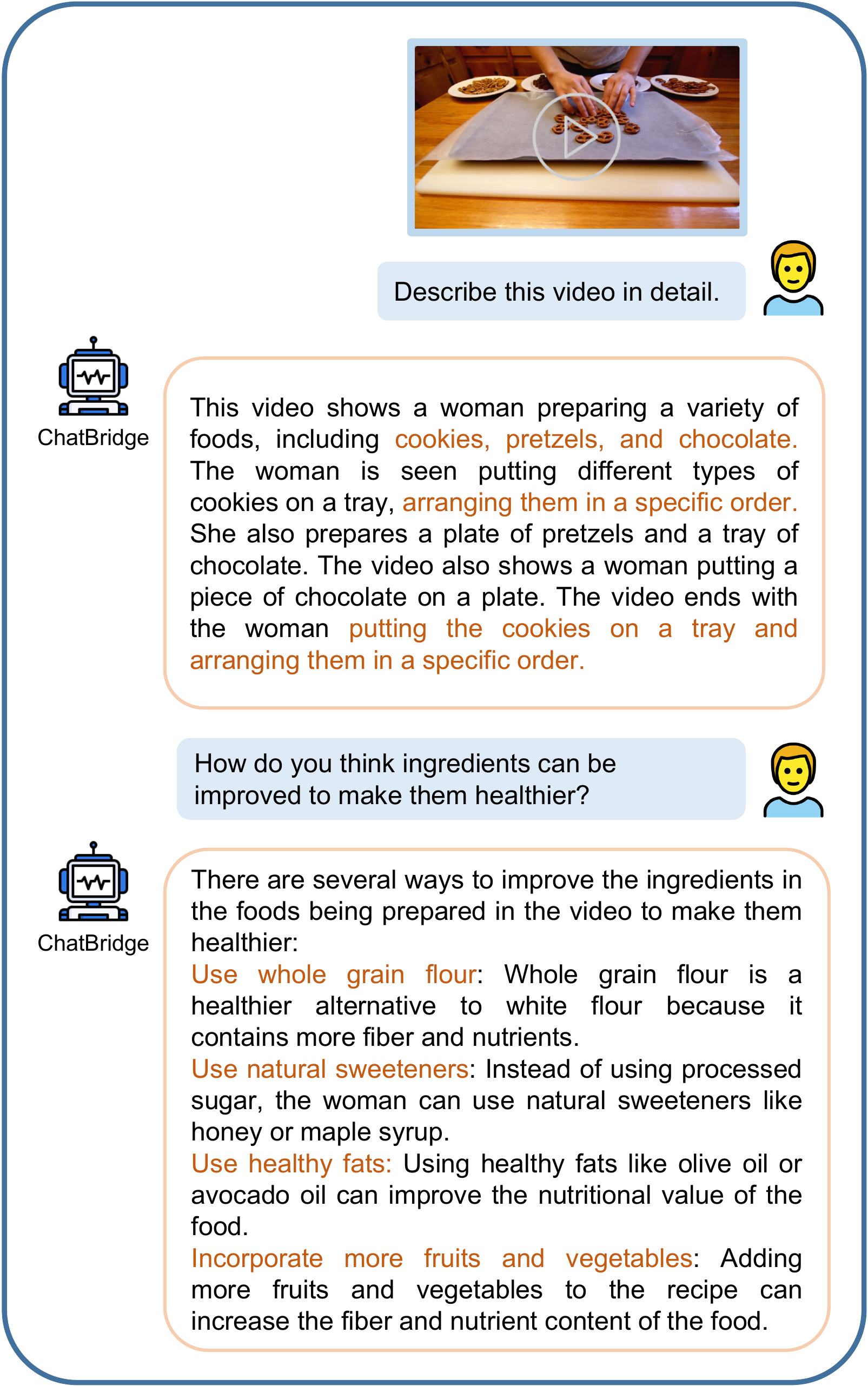}
\caption{Summarizing and extending the video content.}
\end{minipage}
\end{figure}

\begin{figure}[htbp]
\centering
\includegraphics[width=\linewidth]{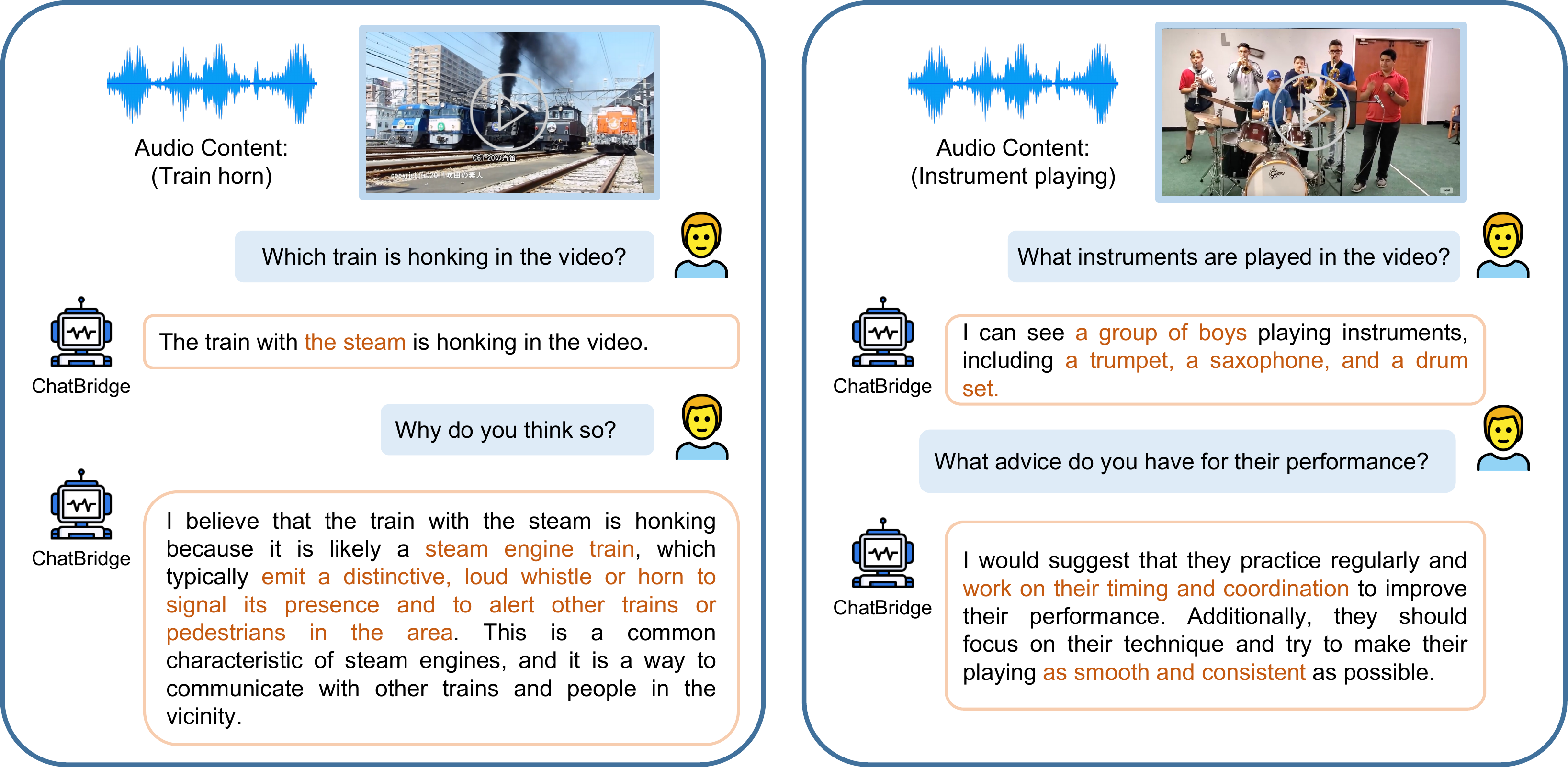}
\caption{Understanding video with its background audio and chatting with human.}
\end{figure}

\begin{figure}[htbp]
\centering
\includegraphics[width=\linewidth]{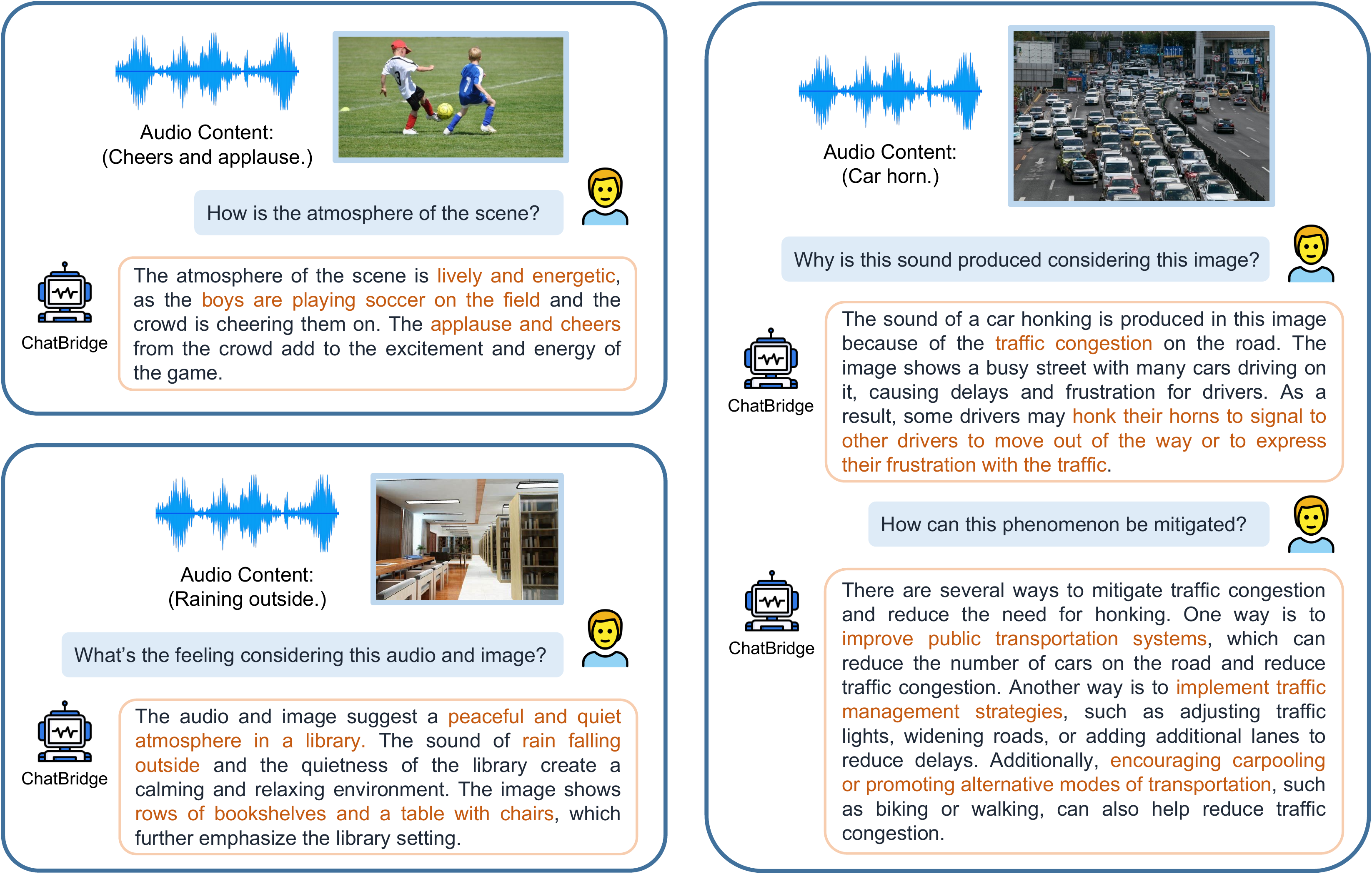}
\caption{Understanding and reasoning about independent image and audio.}
\end{figure}

\begin{figure}[htbp]
\centering
\includegraphics[width=\linewidth]{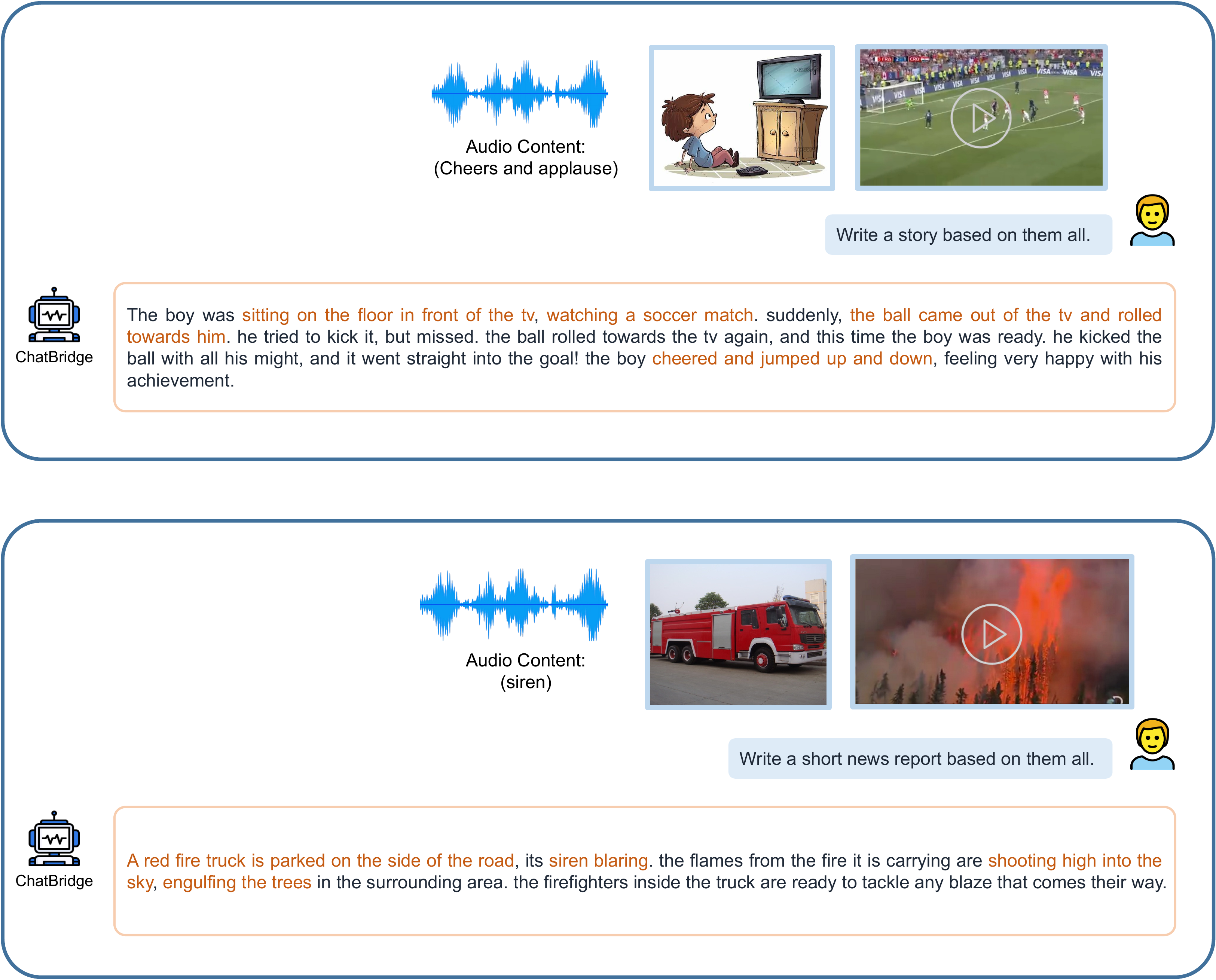}
\caption{After completing the first-stage training, which involves multimodal alignment, the ChatBridge model is already capable of comprehending and analyzing multimodal input where all image, video, and audio modalities are present.}
\end{figure}

\begin{figure}[htbp]
\centering
\includegraphics[width=\linewidth]{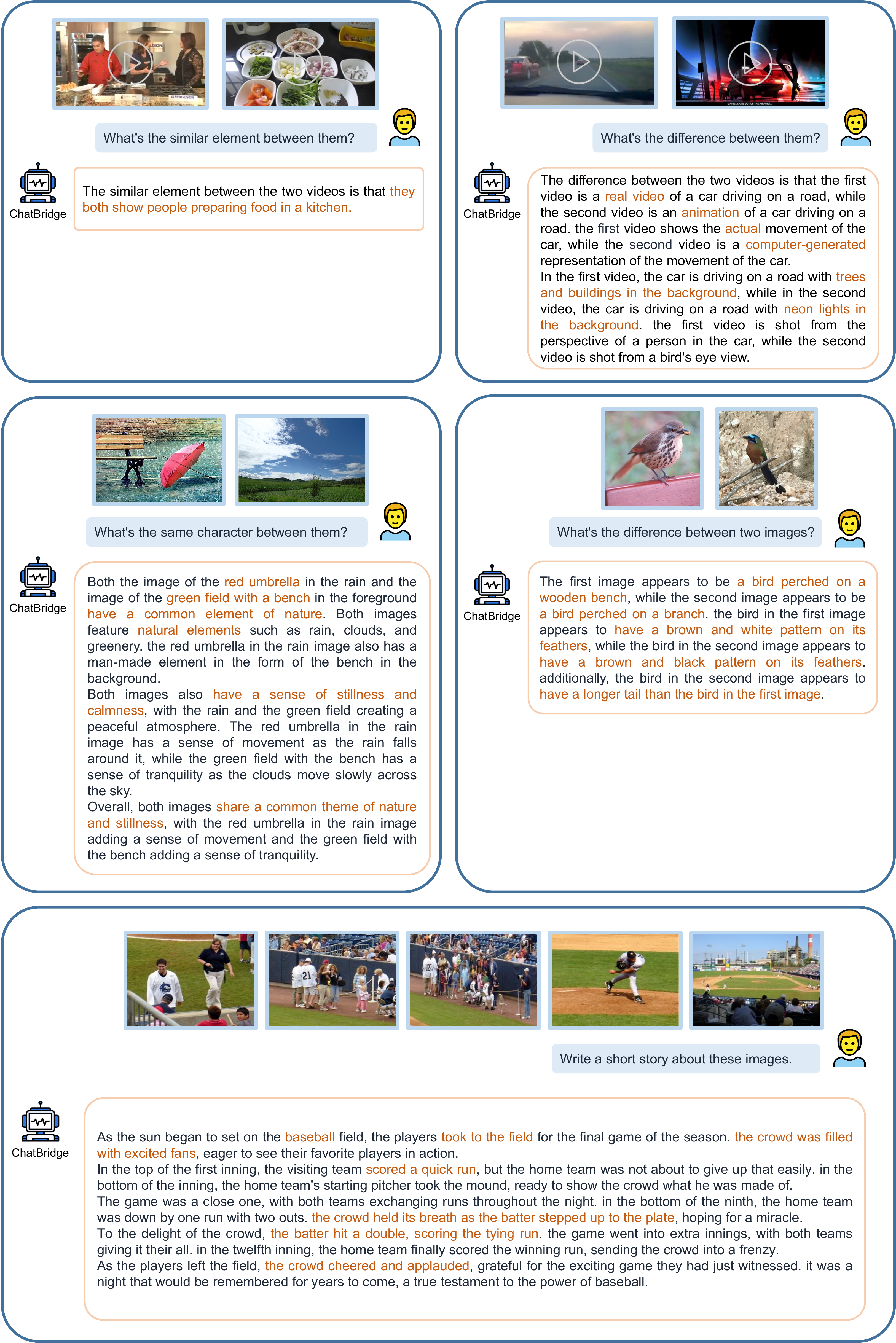}
\caption{ChatBridge model after the first-stage training, \ie multimodal alignment, already has the intrinsic ability to deal with input of multiple images or videos. It can write a story with images, and find the similarities and differences among multiple inputs.}
\end{figure}

\end{document}